\documentclass[sigconf,Zihui Wang,]{acmart}
\makeatletter
\def\@ACM@checkaffil{
    \if@ACM@instpresent\else
    \ClassWarningNoLine{\@classname}{No institution present for an affiliation}%
    \fi
    \if@ACM@citypresent\else
    \ClassWarningNoLine{\@classname}{No city present for an affiliation}%
    \fi
    \if@ACM@countrypresent\else
        \ClassWarningNoLine{\@classname}{No country present for an affiliation}%
    \fi
}
\makeatother

\AtBeginDocument{%
  \providecommand\BibTeX{{%
    \normalfont B\kern-0.5em{\scshape i\kern-0.25em b}\kern-0.8em\TeX}}}

\setcopyright{acmlicensed}
\copyrightyear{2018}
\acmYear{2018}
\acmDOI{XXXXXXX.XXXXXXX}

\acmConference[Conference acronym 'XX]{Make sure to enter the correct
  conference title from your rights confirmation emai}{June 03--05,
  2018}{Woodstock, NY}
\acmISBN{978-1-4503-XXXX-X/18/06}

\usepackage{pifont}
\usepackage{algorithm}

\usepackage{graphicx}
\usepackage{algorithmicx}
\usepackage{algpseudocode}
\usepackage{graphicx}
\usepackage{amsmath}
\usepackage{amsthm}
\usepackage{booktabs}

\usepackage{amssymb}
\usepackage{array} 
\usepackage{pifont}
\usepackage{color}
\usepackage{indentfirst}
\newtheorem{definition}{Definition}
\newtheorem{theorem}{Theorem}
\newtheorem{assumption}{Assumption}
\newcounter{lemma}

\begin{document}

\title{FedSAC: Dynamic Submodel Allocation for Collaborative Fairness \\ in Federated Learning}

\author{Zihui Wang}
\authornote{Both authors contributed equally to this research}
\email{wangziwei@stu.xmu.edu.cn}
\affiliation{
  \institution{Fujian Key Laboratory of Sensing and Computing for Smart Cities, School of Informatics, Xiamen University}
}

\author{Zheng Wang}
\authornotemark[1]
\email{zwang@stu.xmu.edu.cn}
\affiliation{
  \institution{Fujian Key Laboratory of Sensing and Computing for Smart Cities, School of Informatics, Xiamen University}
}

\author{Lingjuan Lyu}
\email{lingjuan.lv@sony.com}
\affiliation{
  \institution{Sony AI}
}

\author{Pengzhao Peng}
\email{pengzhaopeng@stu.xmu.edu.cn}
\affiliation{
  \institution{Fujian Key Laboratory of Sensing and Computing for Smart Cities, School of Informatics, Xiamen University}
}

\author{Zhicheng Yang}
\email{zcyang@stu.xmu.edu.cn}
\affiliation{
  \institution{Fujian Key Laboratory of Sensing and Computing for Smart Cities, School of Informatics, Xiamen University}
}

\author{Chenglu Wen}
\email{clwen@xmu.edu.cn}
\affiliation{
  \institution{Fujian Key Laboratory of Sensing and Computing for Smart Cities, School of Informatics, Xiamen University}
}

\author{Rongshan Yu}
\email{rsyu@xmu.edu.cn }
\affiliation{
  \institution{Fujian Key Laboratory of Sensing and Computing for Smart Cities, School of Informatics, Xiamen University}
}

\author{Cheng Wang}
\email{cwang@xmu.edu.cn}
\affiliation{
  \institution{Fujian Key Laboratory of Sensing and Computing for Smart Cities, School of Informatics, Xiamen University}
}

\author{Xiaoliang Fan}
\email{fanxiaoliang@xmu.edu.cn }
\authornote{Corresponding Author}
\affiliation{
  \institution{Fujian Key Laboratory of Sensing and Computing for Smart Cities, School of Informatics, Xiamen University}
}


\begin{abstract}
\setlength{\parindent}{0em}

Collaborative fairness stands as an essential element in federated learning to encourage client participation by equitably distributing rewards based on individual contributions. Existing methods primarily focus on adjusting gradient allocations among clients to achieve collaborative fairness. However, they frequently overlook crucial factors such as maintaining consistency across local models and catering to the diverse requirements of high-contributing clients. This oversight inevitably decreases both fairness and model accuracy in practice. To address these issues, we propose FedSAC, a novel Federated learning framework with dynamic Submodel Allocation for Collaborative fairness, backed by a theoretical convergence guarantee. First, we present the concept of "\textit{bounded collaborative fairness (BCF)}", which ensures fairness by tailoring rewards to individual clients based on their contributions. 
Second, to implement the BCF, we design a \textit{submodel allocation module} with a theoretical guarantee of fairness. This module incentivizes high-contributing clients with high-performance submodels containing a diverse range of crucial neurons, thereby preserving consistency across local models. Third, we further develop a \textit{dynamic aggregation module} to adaptively aggregate submodels, ensuring the equitable treatment of low-frequency neurons and consequently enhancing overall model accuracy. Extensive experiments conducted on three public benchmarks demonstrate that FedSAC outperforms all baseline methods in both fairness and model accuracy. We see this work as a significant step towards incentivizing broader client participation in federated learning. The source code is available at https://github.com/wangzihuixmu/FedSAC.

\end{abstract}





\maketitle
\section{Introduction}
\label{sec:intro}
\setlength{\parindent}{0pt}Federated Learning (FL) empowers multiple data owners to collectively train a global model while preserving the privacy of their individual training data~\cite{Xu_2023_CVPR11,Mcmahan2017, yan2023criticalfl}. 
Early FL frameworks~\cite{li2020federated,tan2023federated,Chen_2023_CVPR22} usually distributed the same model to all clients without considering their distinct contributions to the model performance, resulting in unfairness to high-contributing clients. \textit{Collaborative fairness} (CF)~\cite{lyu2020collaborative} stands as an essential element in federated learning to motivate client engagement by ensuring impartial reward distribution tied directly to individual contributions.\par

\begin{figure}[tb]
    \centering
    \includegraphics[scale=0.48]{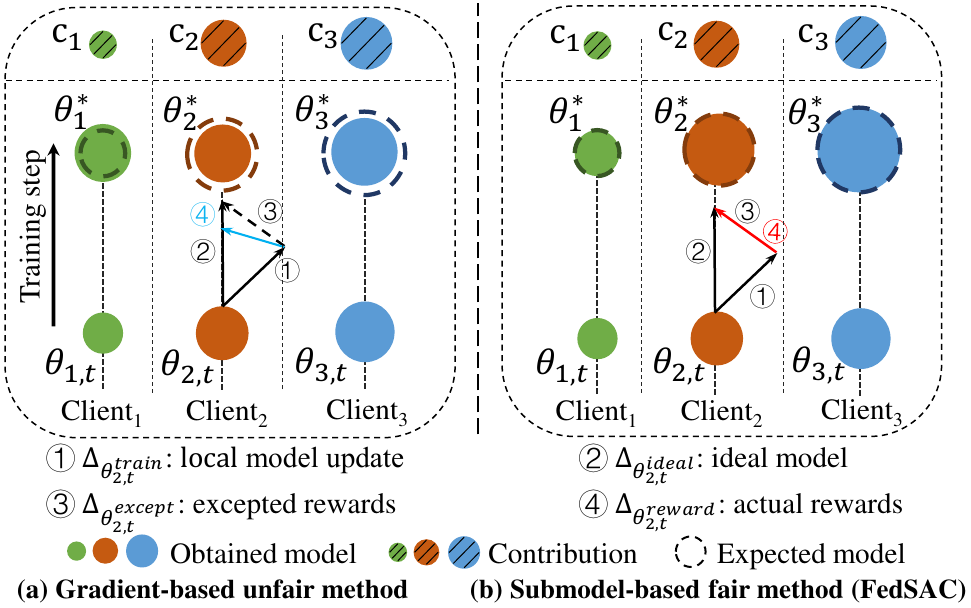}
    \caption{\textbf{Problem illustration of collaborative fairness in FL.} \textbf{(a)} Conventional gradients-based methods will result in poor fairness and model accuracy. For example, it is unfair that obtained models of $Client_2$ and $Client_1$ are equivalent neglecting the inferior contribution ($c_1$) of $Client_1$. Plus, the inconsistency in local models results in that obtained models of $Client_2$ and $Client_3$ are worse than expected ($\theta_i^*$). \textbf{(b)} Our proposed FedSAC allocates sufficient submodels to each client by ensuring a comprehensive balance between fairness and model accuracy. For example, FedSAC ensures that obtained models of all clients (i.e., $Client_1$, $Client_2$ and $Client_3$) are in accordance with their contributions respectively. In addition, FedSAC guarantees the alignment of \ding{174} and \ding{175} of $Client_2$ during the training process, thereby enabling all three clients to obtain their expected models ($\theta_i^*$).}
	\label{fig.1}
\end{figure}

\setlength{\parindent}{1.5em} More recently, several gradient-based methods were proposed to enhance CF~\cite{lyu2020collaborative,xu2020reputation,xu2021gradient, wang2024fedave} (i.e., rewarding clients with corresponding model quality according to their contributions) in FL. They distribute a larger quantity of gradients to higher-contributing clients than the lower ones as rewards and quantify the degree of fairness by Pearson Correlation Coefficient $\rho$. 
However, for achieving CF, existing gradient-based methods have ~\textbf{two major limitations. On one hand}, the conventional definition of CF doesn't adequately distinguish in reward distribution among clients, resulting in a persistent unfairness for high-contributing clients. In Figure~\ref{fig.1} (a), suppose the contributions of three clients are $c=[1,9,11]$, and their rewards are $\theta^*=[99,99.2,99.3]$ corresponding. Through the definition of CF by CGSV~\cite{xu2021gradient}, the fairness is calculated as $\gamma$=98.97, but there exists an underlying unfairness towards $Client_2$ and $Client_3$ because $Client_1$ with an inferior contribution is over-rewarded.
~\textbf{On the other hand}, conventional gradient-based methods~\cite{lyu2020collaborative,xu2020reputation,xu2021gradient, wang2024fedave} are ineffective because the inconsistency of local models updated by variable gradients might lead to significant degradation of overall model performance. In Figure~\ref{fig.1} (a), the local models of clients in round $t$ ($\theta_{1,t}$, $\theta_{2,t}$, and $\theta_{3,t}$) exhibit notable differences (i.e., the larger the circle, the higher the accuracy). Consequently, the gradients uploaded by individual clients may not be the optimal for others, creating a misalignment between obtained rewards $\Delta_{\theta_{i,t}^{reward}}$ (i.e., the rewards ultimately obtained by the clients) and expected rewards $\Delta_{\theta_{i,t}^{exceppt}}$ (i.e., the rewards that the clients ultimately expected) for each client. 


To address the aforementioned challenges, we propose a novel \textit{\underline{Fed}erated learning framework with dynamic \underline{S}ubmodel \underline{A}llocation for bounded \underline{C}ollaborative fairness} (\textbf{FedSAC}), supported by a theory of convergence while achieving competitive model accuracy. 
\textbf{First}, our approach introduces the concept of "\textit{bounded collaborative fairness} (BCF) (refer to $Definition$ \ref{definition1})", which ensures fairness by integrating a differentiated range of rewards allocated to each client. \textbf{Second}, \textit{the submodel allocation module} with a theoretical fairness guarantee, is designed to assign relevant submodels (i.e., results of the aggregated model dropout) to individual clients based on their contributions. Specifically, these submodels encompass a diverse array of essential neurons for effective training.
\textbf{Third}, \textit{the dynamic aggregation module} is implemented as a weight realignment mechanism by treating low-frequency neurons equally, which further improves the overall performance of the global model. Extensive experiments on three public benchmarks show that the proposed FedSAC outperforms all baseline methods in terms of collaborative fairness and model accuracy.\par

\setlength\parindent{1em} The contributions of this work are summarized:
\begin{itemize}
    \item We propose FedSAC, a novel federated learning framework with a convergence guarantee, introducing a new concept of \textit{bounded collaborative fairness (BCF)}. 
    To the best of our knowledge, this is the first approach that allocates submodels equitably for collaborative fairness in FL.
    \item We implement the concept of BCF through two modules. First, \textit{submodel allocation module} prioritizes high-contributing clients by rewarding them with high-performance submodels under a theoretical guarantee. Second, \textit{dynamic aggregation module} merges submodels by paying equitable attention to low-frequency neurons to be aggregated.
    \item We conduct extensive experiments on three benchmarks with various settings and demonstrate that FedSAC ourperforms all baselines in both fairness and model accuracy.
\end{itemize}

\section{Related Works}
\label{sec:Related Works}

\setlength{\parindent}{0pt} Recent research has shown that distributing different rewards based on clients' contributions can significantly impact the FL systems~\cite{yang2019federated,lyu2020democratise,richardson2020budget}. The incentive mechanisms can motivate clients to contribute high-quality data and promote collaboration~\cite{xu2021gradient,shi2021survey,khramtsova2020federated}. We outline three types of rewards that can be adopted to achieve CF in FL.\par 


\setlength\parindent{1.5em} \textbf{Money-based reward.} Several studies focus on the mechanism that rewards clients monetary based on their contributions.~\cite{zhang2021incentive} 
proposes a reputation-based and reverse auction theory mechanism to reward clients with a limited budget.~\cite{yu2020fairness} shows a scheme that dynamically allocates budgets to clients in a context-aware manner by jointly maximizing the collective utility. While monetary rewards can be a natural and effective way to incentivize clients in FL,  there exist challenges in maintaining a  balance between the value of model quality and money~\cite{agarwal2019marketplace,zhan2021survey}.\par

\setlength\parindent{1.5em} \textbf{Data-based reward.} Early studies have explored the fairness of rewarding different data sizes based on their contributions. ~\cite{sim2020collaborative} 
evaluates clients' contributions by aggregating the training data, and reward them with the corresponding models.~\cite{tay2022incentivizing} trains a generative model through the local data of all clients and provides more synthetic data to those datasets closely aligned with the real data distribution.
However, most of existing data-based reward methods rely on the centralized aggregation of all the data during training, making them difficult to be applied in the FL scenarios.\par 

\setlength\parindent{1.5em} \textbf{Gradient-based reward.} Recent collaborative fairness (CF) works aim to reward high-contributing clients with optimal models. CFFL~\cite{lyu2020collaborative} allocates different gradient numbers based on local accuracy in the validation set and data sizes. CGSV~\cite{xu2021gradient} rewards more gradients to clients whose local gradients are more similar to the global gradients. FedAVE~\cite{wang2024fedave} assigns more gradients to clients whose data distribution information is more similar to the ideal dataset. However, existing rewards systems lack sufficient differentiation, resulting in an ongoing unfairness for high-contributing clients, which might degrade the fairness and model accuracy. Different from these methods, we propose a novel framework to achieve BCF (refer to $Definition$ \ref{definition1}) by allocating a differential range of rewards to clients.

\section{Preliminary}

\setlength{\parindent}{0pt} FL system consists of a server and multiple clients, aiming to minimize the weighted average of all clients' local objectives by optimizing a global model~\cite{Mcmahan2017,huang2021personalized,zhang2023fedala,Li_2023_CVPR2121}. First, the server broadcasts a model to the clients at random. Second, after training several rounds locally, the server aggregates these different trained models into a new global model. Finally, the aggregated model will be sent to the clients for further local training. The aforementioned process is repeated multiple times until the global model converges~\cite{Mendieta_2022_CVPR32131,Gao_2022_CVPR32132131, wang2023theoretical}. In this setup, the goal of FL framework is defined as:
\begin{equation}
	\mathop{min}\limits_{\theta}F(\theta) := \sum\limits_{i=1}^{N}p_{i}F_{i}(\theta),
\end{equation}
where $\theta$ denotes the global model, $N$ represents the number of clients, and $p_i=\frac{n_i}{n}$, $n=\sum_{k=1}^{N}n_k$. $F_i(\theta)$ is the loss on client $i$ using model parameters $\theta$, i.e., $F_i(\theta)=\frac{1}{n_i}\sum_{\xi_i\sim{D_i}}{l(\theta, \xi_i)}$, where $D_i$ represents the local dataset of client $i$, and $n_i$ denotes the data size of $D_i$. To achieve this goal as effectively as possible, FedAvg~\cite{Mcmahan2017} samples a subset $S_t$ of $i$ clients uniformly, $0<i
\leq N$, to train the global model and aggregate the locally trained models by utilizing the data size ratio $p_i$ as the weight of client $i$. Although FedAvg is proven to be effective in minimizing the objective successfully, 
it may be unfair to high-quality clients since the system distributes the same rewards to all clients regardless of their contributions~\cite{lyu2020collaborative,xu2021gradient}. 


\subsection{Problem Formulation}
\label{sec3.1}
The standard FL framework allocates the same model to all clients regardless of their contributions~\cite{Mcmahan2017,sun2021partialfed,Huang_2022_CVPR3123,qin2023fedapen}, dampening the motivation of high-quality clients to join FL~\cite{sim2020collaborative,wang2020principled}. Collaborative fairness in FL aims to reward high-contributing clients with high-quality models. The existing works~\cite{lyu2020collaborative,xu2021gradient,wang2024fedave} assess the fairness with the Pearson Correlation Coefficient, $\rho(c;\theta^*)$, where $c$ and $\theta^*$ represent the contributions and rewards of clients, respectively. However, the definition simply considers the relationship between the contributions and rewards of clients, which may lead to insufficient incentives for high-contributing clients. For example in Figure~\ref{fig.1} (a), suppose the contributions of $Client_1$, $Client_2$, and $Client_3$ are $c_i=[1,9,11]$ and their rewards are $\theta_i^*=[99,99.2,99.3]$ correspondingly. Through the definition of CF, the fairness is calculated as $\gamma$=98.97, but there exists an underlying unfairness towards $Client_2$ and $Client_3$ because $Client_1$ with an inferior contribution is 
over-rewarded.\par
\setlength\parindent{1.5em} To address this issue, we propose \textit{Bounded Collaborative Fairness} (BCF) to tackle the issue of insufficient incentives for the high-contributing client. BCF could ensure $c_1<\theta_1^*<\frac{(c_1 + \theta_3^*)}{2}$ and then quantitative fairness with $\rho(c;\theta^*)$. The rationale behind the formula in $Definition$ \ref{definition1} aims to amplify significant distinctions in rewards. The formula's left side ensures that clients' rewards exceed their contributions, while the right side prevents excessive rewards for clients with low contributions.
\begin{definition}[\textbf{Bounded Collaborative Fairness}]
    \label{definition1}
    The contributions ($c$) and the rewards ($\theta^*$) of clients are calculated by the performance of their standalone models (train without collaboration) and the final models obtained after collaboration, respectively. Based on client's obtained rewards $c_i<\theta_i^*<\frac{(c_i + max(\theta^*))}{2}$, the quantitative fairness can be computed by $\gamma$ := 100 $\times$ $\rho$($c$, $\theta^*$) where $\rho$() is the Pearson Correlation Coefficient. The larger $\gamma$, the better the fairness of the framework.
\end{definition}


\section{The Proposed FedSAC}
\begin{figure*}[ht]
	\centering
	\includegraphics[width=0.9\textwidth]{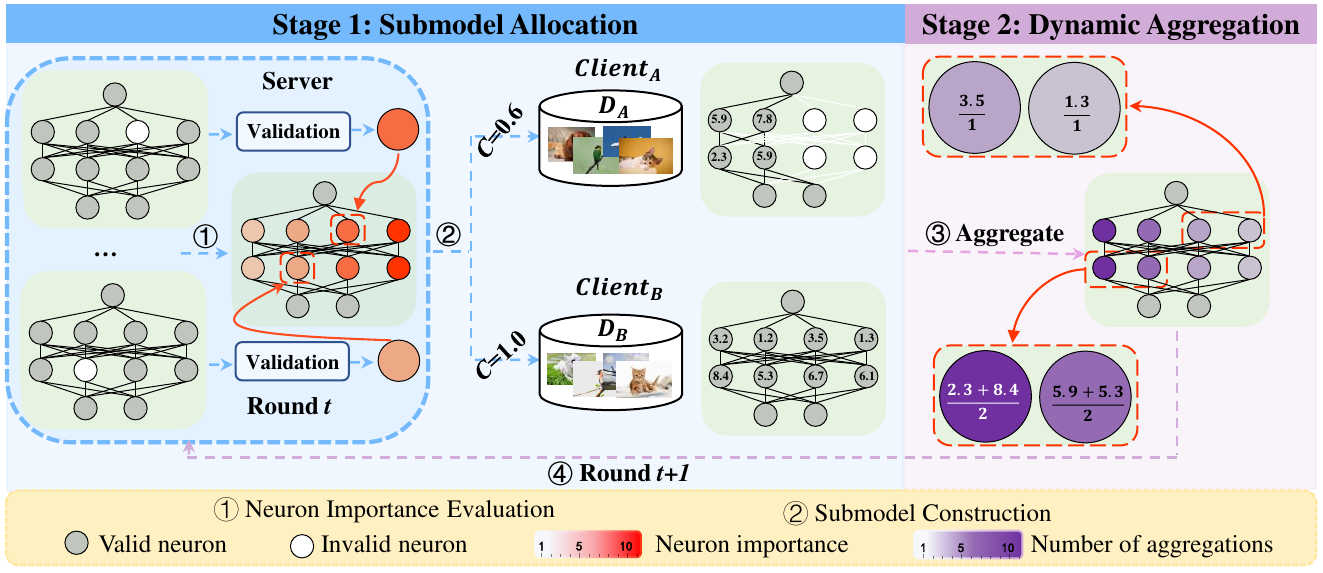}
	\caption{~\textbf{The overall framework of FedSAC} that achieves bounded collaborative fairness by maintaining consistency across local models. FedSAC consists of two module: 1) ~\textit{submodel allocation module} conducts neuron importance evaluation  and submodel construction to reward high-contributing clients with high-performance submodels, thus ensuring consistency in local models; 2) ~\textit{dynamic aggregation module} treats those low-frequency neurons equally, which further refines the performance of the global model.
}
	\label{fig.2}
\end{figure*}

\setlength{\parindent}{0pt} In this section, we will introduce the details of proposed FedSAC, a method that ensures both BCF
and consistency in local models for each client. The architecture of FedSAC is shown in Figure~\ref{fig.2}. The pseudo codes for FedSAC are provided in Algorithm 1.
First, we introduce the submodel allocation module in Section~\ref{sec.4.1}. Second, we present the dynamic aggregation module in Section~\ref{sec.4.2}. Third, we proposed the fairness guarantee theory in Section~\ref{sec.4.3} to prove that this submodel allocation strategy can achieve collaborative fairness. Fourth, we conducted a convergence analysis on FedSAC and demonstrated its convergence in Section~\ref{sec.4.4}. In addition, we analyzed the time complexity and communication costs of FedSAC in Section~\ref{sec.4.4}. Finally, we discussed limitations in Section~\ref{sec.4.5}. 

\par

\subsection{Submodel Allocation Module}
\label{sec.4.1}

\setlength{\parindent}{0pt} A naive approach achieving bounded collaborative fairness involves allocating distinct submodels to each client based on their respective contributions~\cite{hong2022efficient}. Unlike previous works such as~\cite{horvath2021fjord,shi2023towards},  there are two primary motivations behind achieving BCF through submodel allocation. First, submodels with appropriate pruning may not match the performance of the global model, enabling clients to receive diverse submodels according to their contributions. Second, despite being subsets of the global model, these submodels exhibit strong mutual validity, meaning that the submodels uploaded by one client are effective for others, facilitating the training of the global model. However, it is still challenging to achieve BCF through submodel-based methods. For one thing, it is crucial to ensure that the majority of neurons are adequately trained to guarantee the optimal performance of the global model. For another, the performance of allocated submodels should align with their respective contributions.\par 
\setlength\parindent{1.5em} To address the aforementioned two challenges, we design a two-step approach for submodel allocation module. First, we evaluate the importance of each neuron within the model to determine their contributions respectively (\textit{neuron importance evaluation in Section~\ref{sec.4.1.1}}). Second, we construct submodels for each client with varying performances based on their contributions, ensuring a diverse array of important neurons is included within each submodel (\textit{submodel construction in Section~\ref{sec.4.1.2}}).

\subsubsection{\textbf{Neuron Importance Evaluation}}\label{sec.4.1.1}
\setlength{\parindent}{0pt} Each neuron within the model holds a unique contribution~\cite{luo2017thinet,molchanov2016pruning,yu2018nisp,molchanov2019importance}. Our intuition is that the constructed submodels 
can yield varied performances. Inspired by Taylor-FO~\cite{molchanov2019importance}, we calculate the neuron importance in the model by measuring the change in loss upon their removal. For instance, a greater increase in loss indicates a more significant contribution by the removed neuron to the model. In Figure~\ref{fig.2}, neurons depicted in a redder shade represent a higher contribution to the model. Essentially, the training objective is to minimize the cross-entropy loss $L_{ce}$:
\begin{equation}
	\min\limits_{\theta}\sum\limits_{i=1}^{N}L_{ce}(x_{i},\theta),
\end{equation}
where $x_i$ denotes the sample, $\theta$ represents the model, and $L_{ce}(x_{i},\theta)$ is the loss function of the classification tasks.\par
\setlength\parindent{1.5em} The neurons in the model have a multitude of model parameters, each of which contributes to the overall performance of the model. The importance of a neuron $I_{n_i}$ of the model can be calculated through the loss increased by removing it:
\begin{equation}
	\label{eq3}
	I_{n_i} = L_{ce}(V, \theta|\theta_{n_i}^{zero}=0) - L_{ce}(V, \theta),
\end{equation}
where $V$ denotes the validation set, which is constructed by evenly selecting 10\% of the data from the original training samples~\cite{lyu2020collaborative}, $\theta_{n_i}^{zero}$ represents that the parameters of the $i$-th neuron in the model are all set to 0.\par

\setlength\parindent{1.5em} To simplify the construction of submodels, we normalize the sum of neuron scores, which represent their importance in the model presented by percentage: 
\begin{equation}
	\label{eq5}
	I_{n_i} = \frac{I_{n_i}}{\sum\nolimits_{n_i\in{S}}I_{n_i}} *100,
\end{equation}
where $S$ represents all neurons of the model. To reduce the training time of the framework, we measure the importance of neurons in the model by Eq. (\ref{eq3}) and Eq. (\ref{eq5}) every 10 epochs. All these operations allow us to efficiently assess the importance of each neuron while limiting excessive computation demands. \vspace{0.1pt}

\subsubsection{\textbf{Submodel Construction}}\label{sec.4.1.2}
\setlength{\parindent}{0pt} In pursuit of fairness, we employ a dynamic allocation system for submodels with varying performances, leveraging clients' reputations derived from their contributions. Our approach incorporates a pruning mechanism tailored to clients' contributions, simplifying the extraction of submodels with different performance levels from the global model. In this scheme, the client $i$'s reputation $r_i$ is expressed as: 
\begin{equation}
\label{eq51}
	r_i = e^{{c_i}*\beta},
\end{equation}

\begin{equation}
\label{eq6}
	r_i =  \frac{r_i}{max(r)} *100,
\end{equation}
where $c_i$ represents client $i$'s contribution, $\beta$ is a hyper-parameter. The reputations of clients are directly proportional to their contributions. The design rationale for Eq. (\ref{eq51}) and Eq. (\ref{eq6}) is to calculate the clients' reputations ($r$), which facilitates the allocation of their submodels fairly. More specifically, our pruning method begins with the most important neuron, ensuring that submodels for low-contribution clients possess a higher parameter count, which is beneficial for training the global model. These actions serve a dual purpose: promoting collaborative fairness while maximizing the overall performance of the global model.
Submodel $\theta_i$ is constructed by neurons with different reputations:
\begin{equation}
	\theta_i = quantity(r_i, \sum\nolimits_{n_i\in{S}}I_{n_i}),
\end{equation}
where $quantity(r_i, \sum\nolimits_{n_i\in{S}}I_{n_i})$ represents the submodel $\theta_i$ when $r_i = \sum\nolimits_{n_i\in{S}}I_{n_i}$, $r_i$ denotes client $i$'s reputation, $\sum\nolimits_{n_i\in{S}}I_{n_i}$ denotes the set importance for different neurons. $S$ represents the positions of all neurons in the model, arranged in ascending order from the least to the most important. This design choice aims to maximize the inclusion of neurons in each submodel, thereby enhancing the performance of the corresponding local model updates. Subsequently, this quantity will be utilized in Eq. (8) to generate the submodel's mask.

\subsection{Dynamic Aggregation Module}\label{sec.4.2}

Next, the server aggregates the locally trained submodels and allocates distinct submodels to clients in the subsequent round. Recent submodel-based methods~\cite{horvath2021fjord,shi2023towards} have aimed to allocate varied submodels containing numerous neurons to clients. However, these approaches might pose a potential risk of compromising overall model performance when integrating low-frequency neurons into the global model. Consequently, employing a direct aggregation method such as FedAvg~\cite{Mcmahan2017} for all neurons becomes inequitable.\par

\setlength\parindent{1.5em} Instead of simply averaging the uploaded submodels, our objective is to optimize the utilization of all neurons within the model. With the sizes of submodels varying across clients, it becomes essential to treat the contribution of each neuron individually during aggregation. To ensure fair treatment of low-frequency neurons, we integrate the frequency of submodel parameter aggregations as weights to dynamically aggregating local models:
\begin{equation}
	mask_i^t = mask(\theta_i^t, \theta_g^{t-1}),
\end{equation}
\begin{equation}
\label{eq10}
	\theta_g^{t+1} = \frac{\sum\nolimits_{i\in{N}}\theta_i^t}{\sum\nolimits_{i\in{N}}mask_i^t},
\end{equation}
where $mask(\theta_i, \theta_g)$ denotes submodel $\theta_i$'s mask (same shape as the submodel $\theta_i$), $\theta_g$ denotes the aggregated model, $N$ denotes the total number of clients. It sets the components of both $\theta_i$ and $\theta_g$ at the same position to 1 and 0 for the rest. The role of the mask function $mask(\theta_i, \theta_g)$ is to calculate the frequency of each model parameter $\theta_i$ selected by the global model $\theta_g$ in round $t$. Later, the mask function $mask(\theta_i, \theta_g)$ will be utilized in Eq. (9) to treat those low-frequency parameters equally by suppressing the weight of high-frequency parameters in the aggregation, which makes each parameter play a fair role during the aggregation phase. For example, the more the frequency of a selected parameter in round $t$, the smaller the weight of the parameter to be aggregated in round $t+1$.
\subsection{Fairness Guarantee}\label{sec.4.3}
\setlength{\parindent}{0pt} In Section~\ref{sec.4.1}, we delved into the fundamental concept underpinning our definition of fairness. This concept centers on rewarding high-contributing clients with high-performance submodels, where a submodel's improved performance correlates with the number of neurons it contains. Consequently, this approach leads to a training loss (i.e., model accuracy) that more closely aligns with the aggregated model. It's important to note that the submodel $\theta_i$ acquired by client $i$ is determined based on its reputation $r_i$ across the entire training process up to iteration $t$.\par
\setlength\parindent{1.5em}Our primary result ensures a notion of fairness under specific conditions concerning the loss function $F$. If client $i$ holds a higher reputation than client $j$ ($r_i \ge r_j$), and the submodel $\theta_i^t$ obtained by client $i$ encompasses the submodel $\theta_j^t$ obtained by client $j$ ($\theta_j^t \in \theta_i^t \in \theta_g^t$). Then, the submodel $\theta_i^t$ obtained by client $i$ will exhibit closer alignment with the aggregated model $\theta_g^t$ in round $t$. Letting $\delta_i^t := ||\theta_g^t - \theta_i^t||$, it's evident that $\delta_i^t \leqslant \delta_j^t$. Consequently, the submodel $\theta_i^t$ obtained by client $i$ will yield a smaller loss function $F(\theta)$ compared to client $j$ in round $t$.

\setlength{\parindent}{0em}\begin{assumption}[\textbf{$L$-smooth F}]
	If $F$ is $L$-smooth, then ${\forall} \theta_i, \theta_j \in \theta$,
\begin{equation}
	F(\theta_i) \leqslant F(\theta_j) + \nabla F(\theta_j)^T(\theta_i - \theta_j) + \frac{L}{2}||\theta_i - \theta_j||^2.
\end{equation}
\label{Assumption1}
\end{assumption}
\vspace{-20pt}
\begin{assumption}[\textbf{$\mu$-strongly convex F}]
	If $F$ is $\mu$-strongly convex, then ${\forall} \theta_i, \theta_j \in \theta$,
\begin{equation}
	F(\theta_i) \ge F(\theta_j) + \nabla F(\theta_j)^T(\theta_i - \theta_j) + \frac{\mu}{2}||\theta_i - \theta_j||^2.
\end{equation}
\label{Assumption2}
\end{assumption}
\vspace{-20pt}
\begin{theorem}[\textbf{Fairness in Training Loss}]
    \textit{Assume Assumptions 1 and 2 hold, FedSAC can guarantee collaborative fairness by rewarding high-contributing clients obtaining high-performance models. Formally speaking, let $\delta_i^t := ||\theta_g^t - \theta_i^t||$. Suppose that $\theta_t$ is close to a stationary point of $F$ for $t\ge T\in Z^+$, and $F()$ is both $L$-smooth and $\mu$-strongly convex with $L\leqslant\mu$. For all $i,j \in N$ in round t, if $r_i \ge r_j$, it follows that $\theta_j^t \in \theta_i^t \in \theta_g^t$, $\delta_i^t \leqslant \delta_j^t$, and therefore $F(\theta_i^t) \leqslant F(\theta_j^t)$.}

\end{theorem}

The proof process is as follows:

\setlength{\parindent}{0pt} From $L$-smoothness (ASSUMPTION~\ref{Assumption1}), we have
\begin{equation}
F(\theta_i^t) \leqslant \underbrace{F(\theta_N^t) + \nabla F(\theta_N^t)^T(\theta_i^t - \theta_N^t) + \frac{L}{2}\delta_{i,t}^2}_{\text{$R_L$}}.
\end{equation}

From $\mu$-strongly convex (ASSUMPTION~\ref{Assumption2}), we have\par
\begin{equation}
F(\theta_j^t) \ge \underbrace{F(\theta_N^t) + \nabla F(\theta_N^t)^T(\theta_j^t - \theta_N^t) + \frac{\mu}{2}\delta_{j,t}^2}_{\text{$R_\mu$}}.
\end{equation}

In order to prove $F(\theta_i^t) \leqslant F(\theta_j^t)$, it suffices to prove $R_L \leqslant R_\mu$ or equivalently $R_L - R_\mu \leqslant 0$.\par

\begin{equation}
R_L - R_\mu =  \underbrace{\nabla F(\theta_N^t)^T(\theta_i^t - \theta_j^t)}_{\text{$R_1$}} + \underbrace{\frac{1}{2}(L\delta_{i,t}^2 - \mu\delta_{j,t}^2)}_{\text{$R_2$}}.
\end{equation}

With  $L \leqslant \mu$ and  $\delta_{i,t} \leqslant \delta_{j,t}$, we have
\begin{equation}
\label{eq16}
R_2 = \frac{1}{2}(L\delta_{i,t}^2 - \mu\delta_{j,t}^2) \leqslant \frac{L}{2}(\delta_{i,t}^2 - \delta_{j,t}^2) \leqslant 0.
\end{equation}

\setlength{\parindent}{0pt}
We define $\theta_N^t$ being close to a stationary point of F by establishing an upper limit on the gradient:\par

\begin{equation}
\label{eq17}
||\nabla F(\theta_N^t) ||\leqslant \frac{L|\delta_{i,t}^2 - \delta_{j,t}^2|}{2||\theta_i^t-\theta_j^t||}.
\end{equation}

\setlength{\parindent}{0pt}We have the following:
\begin{equation}
\begin{split}
|R_1| &\triangleq | \nabla F(\theta_N^t)^T(\theta_i^t - \theta_j^t)| \\
&\leqslant ||\nabla F(\theta_N^t)|| \times ||(\theta_i^t-\theta_j^t)|| \\
&\leqslant \frac{L|\delta_{i,t}^2 - \delta_{j,t}^2|}{2}\\
&\leqslant |R_2|,
\end{split}
\end{equation}

\setlength{\parindent}{0pt} where the first inequality is derived from the Cauchy-Schwarz, the second inequality is by substituting the
aforementioned upper limit (refer to Eq.~(\ref{eq17})), and the last inequality (line 1209) emerges from taking the absolute values of two negative values (refer to Eq.~(\ref{eq16})).\par

\setlength{\parindent}{1.5em}
Finally, given that $|R_1| \leqslant |R_2|$ and $R_2 \leqslant 0$, we derive $	R_1 + R_2 \leqslant 0$. Therefore, it follows that $R_L - R_\mu \triangleq R_1 + R_2 \leqslant 0$, which subsequently implies $F(\theta_i^t) \leqslant F(\theta_j^t)$.

\begin{algorithm}[tb]
\caption{FedSAC}  
\begin{algorithmic}[1]
    \renewcommand{\algorithmicrequire}{ \textbf{Input:}}    
    \renewcommand{\algorithmicensure}{ \textbf{Output:}}
	
    \Require
    The global model $\theta_g$, the local submodel $\theta_i$, neurons $i$'s importance $I_{n_i}$, the number of local update steps $E$, learning rate $\eta_t$, number of clients $N$, hyper-parameter $\beta$, client's contribution $c$
	
    \State Initialize the global model parameters $\theta_g^0$
    \For {round $t=0,1,...,T-1$}
    \State Compute $I_{n_i}^t$ (\ref{eq3}) and (\ref{eq5}) of $\theta_g^t$
    \State Calculate the reputation $r_i$ of client $k$: 	$r_i = \frac{e^{{c_i}*\beta}}{max(e^{{c}*\beta})}*100$
    \State Calculate Submodels $\theta_i^t$ of clients $i$ in round $t$: 	$\theta_i^t = quantity(r_i, \sum\nolimits_{n_i\in{S}}I_{n_i}^t)$
    \For {each client $i\in{N}$}
    \For {each iteration $j=0,1,...,E-1$}
    \State $\theta_{i,j+1}^t\gets{\theta_{i,j}^t} -$ $\eta_t\nabla{F_i(\theta_{i,j}^t)}$ 
    \EndFor
    \EndFor
    \State Submodel $i$'s mask in round $t$: $mask_i^t = mask(\theta_i^t, \theta_g^{t-1})$
    \State The server aggregates the received submodels: $\theta_g^{t+1} = \frac{\sum\nolimits_{i\in{N}}\theta_i^t}{\sum\nolimits_{i\in{N}}mask_i^t}$ 
    
    \EndFor
\end{algorithmic}  
\end{algorithm}
\subsection{Convergence Analysis}
In this section, we delve into the convergence analysis of the proposed FedSAC. To guarantee convergence to the global optimum, we make the assumption that each neuron in the aggregated model is equally allocated over $T$ rounds. Consequently, the anticipated weight of the allocated submodel $\theta_i$ contracts towards the aggregate model $\theta_g$, i.e., $\theta_i^{t+1}=p_i\theta_g^{t}$. Here, $p_i$ ($0\leqslant p_i \leqslant 1$) denotes the long-term expectation of the size ratio between the submodel $i$ and the aggregate model obtained in multiple iterations. 
At this stage, Eq. (\ref{eq10}) can be expressed as the aggregation of each submodel $\theta_i$ divided by its respective $p_i$, i.e., $\theta_g^{t+1} = \sum_{i=1}^{N}\frac{\theta_i^{t+1}}{p_i}$.
We present THEOREM~\ref{theorem2} below, which demonstrates that FedSAC enables the convergence of the aggregation model. Assumptions~\ref{Assumption3} and~\ref{Assumption4} are derived from the works~\cite{zhang2012communication, stich2018local, stich2018sparsified, yu2019parallel}.\\

\vspace{-10pt}
\begin{assumption}
 \label{Assumption3}
 Let $\xi_i^t$ denote samples uniformly from the local data of the $i$-th device at random. It is asserted that the variance of stochastic gradients within each device remains constrained:
\begin{equation}
E\|\nabla F_i(\theta_i^t, \xi_i^t) - \nabla F_i(\theta_i^t)\| \leqslant \sigma_i^2,
\end{equation}
\end{assumption}

\begin{assumption}
 \label{Assumption4}
    The expected squared norm of stochastic gradients is uniformly constrained:
    \begin{equation}
        E\|\nabla F_i(\theta_i^t, \xi_i^t)\| \leqslant G^2,
    \end{equation}
where $i \in \{1, 2, ..., $N$\}$ and $t \in \{1, 2, ..., T-1 \}$.
\end{assumption}

\begin{assumption}
\label{Assumption5}
Each neuron in the aggregation model is assigned the same number of times after $T$ rounds. Therefore, the expected weight of the allocated submodel $\theta_i$ is a contraction of the aggregate model $\theta_g$, i.e., $\theta_i^{t+1}=p_i\theta_g^t$.  Here, $p_i$ ($0\leqslant p_i \leqslant 1$) denotes the long-term expectation of the size ratio between the submodel $i$ and the aggregate model obtained in multiple iterations.
\end{assumption}

\begin{theorem}[\textbf{Asymptotic convergence}]
\label{theorem2}
Given that Assumptions~\ref{Assumption1} to~\ref{Assumption5} hold and $L$, $\mu$, $\sigma_i$, $G$, $p$ be defined therein. Choose $\kappa=\frac{2}{\mu}$, $\gamma=max\{8\frac{L}{\mu}, E\}-1$ and the learning rate $\eta_t = \frac{2}{\mu(\gamma+t)}$. Then FedSAC satisfies $ E[F(\bar{\theta}_T)] - F^* \leqslant [\frac{L}{\gamma+T}(\frac{2B}{\mu^2}+ \frac{\gamma+1}{2}\triangle_1)]$.

\end{theorem}
The proof is shown in Appendix A.\\
\subsection{Complexity and Communication Cost Analysis}\label{sec.4.4}
We further analyze the time complexity and communication costs of FedSAC as follows.\par
\setlength{\parindent}{1.5em}\textbf{For the time complexity}, the primary computational demand in FedSAC stems from evaluating neuron importance, as defined in Eq. (\ref{eq3}) and Eq. (\ref{eq5})). The time complexity for this evaluation is $O(M)$, where $M$ denotes the total number of neurons across the hidden layers of the global model.\par
\setlength{\parindent}{1.5em}\textbf{For the communication cost}, FedSAC mitigates the introduction of additional communication overhead by conducting neuron importance evaluation solely on the server. This approach effectively eliminates the necessity for client-server communication, thereby enhancing overall efficiency.
Moreover, it displays a communication complexity of O(d*m) per round, as outlined in~\cite{horvath2021fjord}, where m$\leqslant$1 denotes the average ratio of submodel parameters to the global model. As a result, FedSAC showcases lower communication complexity compared to all baseline methods in cross-silo FL scenarios~\cite{ogier2022flamby}.  More details about the communication cost experiments are put in Appendix E.\par

\subsection{Limitations}\label{sec.4.5}

\setlength{\parindent}{0pt} In Table \ref{tabel1} and Table \ref{table2}, we conduct extensive experiments on various datasets and observe that FedSAC could exhibit a distinct advantage over all baseline methods in terms of both fairness and model accuracy. Nevertheless, the sufficient evaluation of neuron importance (Section~\ref{sec.4.1.1}) within the submodel allocation module imposes an additional computational burden. This problem may be amplified for large models. Despite this challenge, we hold a strong conviction that the substantial enhancements in both fairness and accuracy achieved through by FedSAC clearly affirm its superiority over baseline methods.

\section{Experiments}

\setlength{\parindent}{0pt} In this section, we conduct comprehensive experiments to answer the following research questions:\par
\setlength{\parindent}{1.5em}\textbf{RQ1.} How does the fairness of our FedSAC compare to various state-of-the-art methods?\par
\setlength{\parindent}{1.5em}\textbf{RQ2.} How does the predictive model performance achieved by our proposed method compare with the state-of-the-art methods on different datasets?\par
\setlength{\parindent}{1.5em}\textbf{RQ3.} How do different components (i.e., submodel allocation module and dynamic aggregation module) affect the results?\par

\subsection{Experimental Settings}\label{sec.3.1}
\setlength{\parindent}{0pt} \noindent\textbf{Datasets and Models.} We evaluate the performance of FedSAC on three commonly used public datasets in collaborative fairness, including Fashion MNIST~\cite{lecun1998gradient}, CIFAR10~\cite{krizhevsky2009learning}, and SVHN~\cite{netzer2011reading}. Following~\cite{lin2020ensemble,li2021fedmask,uddin2020mutual}, we employ a feedforward neural network with two hidden layers for all datasets.\par
\setlength\parindent{1.5em}\indent \noindent\textbf{Data splits.} We construct five heterogeneous scenarios by varying the size and the class numbers of the dataset. For \textbf{imbalanced dataset sizes} (POW)~\cite{lyu2020collaborative,xu2021gradient}, we randomly divide the total dataset into various data sizes for each client by using a power law.  For CIFAR10, we partition the data set of size 20000 among 10 clients. The clients with more extensive data sizes are expected to achieve better prediction performance. For \textbf{imbalanced class numbers} (CLA) ~\cite{lyu2020collaborative,xu2021gradient}, we change the number of classes and keep them have the same amount of data. For CIFAR10 with 5 clients, clients 1, 2, 3, 4, 5 own local training data with 1, 3, 5, 7, 10 classes respectively. For \textbf{imbalanced data size and class numbers} (DIR), we provide clients with various data sizes and classes by the Dirichlet distribution function~\cite{chen2022gear,yurochkin2019bayesian,gao2022feddc,yu2022tct}. Specifically, we sample $p_{i}^l \sim DIR(\alpha)$ and assign a $p_{i}^l$ percentage of the data of class $l$ to client $i$, where $DIR(\alpha)$ is the  Dirichlet distribution with a parameter $\alpha$. More details on the varying numbers of clients used in the experiment are put in Appendix C.\par

\begin{table*}
    \centering
    \resizebox{1.0\linewidth}{!}{
    \renewcommand\arraystretch{1.3}
        \large
        \begin{tabular}{@{}l|c|ccccc|ccccc|ccccc@{}}
            \toprule
            \quad  Dataset        & \multicolumn{5}{c|}{CIFAR10} &\multicolumn{5}{c|}{SVHN} &\multicolumn{5}{c}{Fashion MNIST}\\ 
             \midrule
            \quad  No. Clients        & \multicolumn{5}{c|}{10} &\multicolumn{5}{c|}{10} &\multicolumn{5}{c}{10}\\ 
             \midrule
		\quad Scene & \multicolumn{1}{c}{POW}            & \multicolumn{1}{c}{CLA}                & \multicolumn{1}{c}{DIR(1.0)}               & \multicolumn{1}{c}{DIR(2.0)}            & \multicolumn{1}{c|}{DIR(3.0)}     
				& \multicolumn{1}{c}{POW}            & \multicolumn{1}{c}{CLA}                  & \multicolumn{1}{c}{DIR(1.0)}               & \multicolumn{1}{c}{DIR(2.0)}            & \multicolumn{1}{c|}{DIR(3.0)} & \multicolumn{1}{c}{POW}            & \multicolumn{1}{c}{CLA}                  & \multicolumn{1}{c}{DIR(1.0)}               & \multicolumn{1}{c}{DIR(2.0)}            & \multicolumn{1}{c}{DIR(3.0)}\\
            \midrule

        \quad 	FedAvg\cite{Mcmahan2017}         & \multicolumn{1}{c}{14.55$\pm$25.4}   & \multicolumn{1}{c}{84.68$\pm$3.5}                       & \multicolumn{1}{c}{10.77$\pm$6.3}              & \multicolumn{1}{c}{15.06$\pm$21.2}              & 
				\multicolumn{1}{c|}{68.93$\pm$8.1}            & \multicolumn{1}{c}{-19.24$\pm$26.7}          & \multicolumn{1}{c}{71.42$\pm$13.0}                      & \multicolumn{1}{c}{58.56$\pm$4.0}              & \multicolumn{1}{c}{50.95$\pm$13.5}              & 
				\multicolumn{1}{c|}{31.43$\pm$19.0}   & \multicolumn{1}{c}{-19.83$\pm$24.1}          & \multicolumn{1}{c}{78.44$\pm$4.3}             & \multicolumn{1}{c}{17.94$\pm$7.3}              & \multicolumn{1}{c}{60.36$\pm$16.3}              & 
				\multicolumn{1}{c}{76.09$\pm$29.5} \\

	\quad q-FFL\cite{li2019fair}           & \multicolumn{1}{c}{34.91$\pm$15.1}          & \multicolumn{1}{c}{\underline{98.74}$\pm$0.5}                       & \multicolumn{1}{c}{\underline{90.89}$\pm$1.5}              & \multicolumn{1}{c}{\underline{89.01}$\pm$1.5}              & 
				\multicolumn{1}{c|}{73.02$\pm$2.5}   & \multicolumn{1}{c}{67.97$\pm$11.6}          & \multicolumn{1}{c}{93.05$\pm$1.0}                & \multicolumn{1}{c}{82.11$\pm$1.8}              & \multicolumn{1}{c}{\underline{86.87}$\pm$4.7}              & 
				\multicolumn{1}{c|}{\underline{89.41}$\pm$2.1}   & \multicolumn{1}{c}{42.69$\pm$9.8}          & \multicolumn{1}{c}{\underline{88.99}$\pm$0.3}  & \multicolumn{1}{c}{73.19$\pm$6.0}              & \multicolumn{1}{c}{83.93$\pm$0.7}              & 
				\multicolumn{1}{c}{79.99$\pm$7.3} \\

	\quad CFFL\cite{lyu2020collaborative}           & \multicolumn{1}{c}{\underline{93.55}$\pm$1.3}          & \multicolumn{1}{c}{89.99$\pm$0.8}                     & \multicolumn{1}{c}{29.90$\pm$3.6}              & \multicolumn{1}{c}{81.82$\pm$0.8}              & 
				\multicolumn{1}{c|}{59.86$\pm$3.0}   & \multicolumn{1}{c}{\underline{96.38}$\pm$1.5}          & \multicolumn{1}{c}{\underline{95.63}$\pm$0.4}                & \multicolumn{1}{c}{46.91$\pm$5.7}              & \multicolumn{1}{c}{38.58$\pm$2.5}              & 
				\multicolumn{1}{c|}{31.35$\pm$8.3}   & \multicolumn{1}{c}{90.94$\pm$0.5}      & \multicolumn{1}{c}{86.50$\pm$0.7}         & \multicolumn{1}{c}{\underline{85.90}$\pm$0.9}              & \multicolumn{1}{c}{85.09$\pm$1.3}              & 
				\multicolumn{1}{c}{71.10$\pm$0.3} \\

	\quad CGSV\cite{xu2021gradient}             & \multicolumn{1}{c}{90.78$\pm$0.6}          & \multicolumn{1}{c}{91.04$\pm$0.8}                & \multicolumn{1}{c}{63.29$\pm$3.6}              & \multicolumn{1}{c}{84.59$\pm$1.6}              & 
				\multicolumn{1}{c|}{84.75$\pm$0.2}   & \multicolumn{1}{c}{90.99$\pm$0.4}       &\multicolumn{1}{c}{87.22$\pm$0.6}        &\multicolumn{1}{c}{72.09$\pm$0.2}              & \multicolumn{1}{c}{72.19$\pm$0.2}              & 
				\multicolumn{1}{c|}{76.31$\pm$0.2}   & \multicolumn{1}{c}{\underline{95.34}$\pm$0.3}          & \multicolumn{1}{c}{73.65$\pm$2.6}               & \multicolumn{1}{c}{82.91$\pm$3.6}              & \multicolumn{1}{c}{82.91$\pm$3.6}              & 
				\multicolumn{1}{c}{\underline{84.95}$\pm$1.3} \\

	\quad FedAVE\cite{wang2024fedave}             & \multicolumn{1}{c}{85.50$\pm$0.8}          & \multicolumn{1}{c}{92.80$\pm$1.2}                & \multicolumn{1}{c}{58.43$\pm$0.9}              & \multicolumn{1}{c}{85.82$\pm$0.7}              & 
				\multicolumn{1}{c|}{\underline{88.61}$\pm$1.3}   & \multicolumn{1}{c}{92.68$\pm$0.4}       &\multicolumn{1}{c}{92.77$\pm$0.8}        &\multicolumn{1}{c}{\underline{82.42}$\pm$1.2}              & \multicolumn{1}{c}{64.83$\pm$1.0}              & 
				\multicolumn{1}{c|}{79.38$\pm$0.8}   & \multicolumn{1}{c}{86.98$\pm$0.5}          & \multicolumn{1}{c}{86.36$\pm$1.1}               & \multicolumn{1}{c}{79.51$\pm$1.3}              & \multicolumn{1}{c}{\underline{87.99}$\pm$0.5}              & 
				\multicolumn{1}{c}{67.62$\pm$0.3} \\

	\quad Ours      & \multicolumn{1}{c}{\textbf{98.80}$\pm$0.2}          & \multicolumn{1}{c}{\textbf{99.06}$\pm$0.3}               & \multicolumn{1}{c}{\textbf{99.14}$\pm$0.6}              & \multicolumn{1}{c}{\textbf{95.73}$\pm$0.5}              & 
				\multicolumn{1}{c|}{\textbf{97.01}$\pm$0.7}   & \multicolumn{1}{c}{\textbf{99.44}$\pm$0.3}          & \multicolumn{1}{c}{\textbf{99.74}$\pm$0.1}              & \multicolumn{1}{c}{\textbf{96.09}$\pm$0.3}              & \multicolumn{1}{c}{\textbf{96.48}$\pm$0.2}              & 
				\multicolumn{1}{c|}{\textbf{98.32}$\pm$0.8}   & \multicolumn{1}{c}{\textbf{96.35}$\pm$0.2}          & \multicolumn{1}{c}{\textbf{98.93}$\pm$0.7}              & \multicolumn{1}{c}{\textbf{99.23}$\pm$0.3}              & \multicolumn{1}{c}{\textbf{97.71}$\pm$0.2}              & 
				\multicolumn{1}{c}{\textbf{98.62}$\pm$0.3} \\
            \bottomrule
        \end{tabular}
        }
    \caption{~\textbf{Comparison results of fairness $\rho\in[-100, 100]$} with state-of-the-art methods on three datasets. The reported results are averaged over 5 runs with different random seeds. (A higher value indicates better fairness. The best average result is marked in bold. The second-best result is underlined. These notes are the same to others.)}
    \label{tabel1}
\end{table*}

\begin{table*}
    \centering
    \resizebox{1.0\linewidth}{!}{
     \renewcommand\arraystretch{1.3}
        \large
        \begin{tabular}{@{}l|c|ccccc|ccccc|ccccc@{}}
            \toprule
            \quad  Dataset        & \multicolumn{5}{c|}{CIFAR10} &\multicolumn{5}{c|}{SVHN} &\multicolumn{5}{c}{Fashion MNIST}\\ 
             \midrule
            \quad  No. Clients        & \multicolumn{5}{c|}{10} &\multicolumn{5}{c|}{10} &\multicolumn{5}{c}{10}\\ 
             \midrule
		\quad Scene & \multicolumn{1}{c}{POW}            & \multicolumn{1}{c}{CLA}                & \multicolumn{1}{c}{DIR(1.0)}               & \multicolumn{1}{c}{DIR(2.0)}            & \multicolumn{1}{c|}{DIR(3.0)}     
				& \multicolumn{1}{c}{POW}            & \multicolumn{1}{c}{CLA}                  & \multicolumn{1}{c}{DIR(1.0)}               & \multicolumn{1}{c}{DIR(2.0)}            & \multicolumn{1}{c|}{DIR(3.0)} & \multicolumn{1}{c}{POW}            & \multicolumn{1}{c}{CLA}                  & \multicolumn{1}{c}{DIR(1.0)}               & \multicolumn{1}{c}{DIR(2.0)}            & \multicolumn{1}{c}{DIR(3.0)}\\
            \midrule
        \quad Standalone & \multicolumn{1}{c}{41.23$\pm$0.1}            & \multicolumn{1}{c}{37.49$\pm$0.2}           & \multicolumn{1}{c}{33.78$\pm$0.2}               & \multicolumn{1}{c}{33.54$\pm$0.1}            & \multicolumn{1}{c|}{31.89$\pm$0.1}     
				& \multicolumn{1}{c}{60.02$\pm$0.2}            & \multicolumn{1}{c}{52.05$\pm$0.3}            & \multicolumn{1}{c}{41.41$\pm$0.2}               & \multicolumn{1}{c}{58.07$\pm$0.2}            & \multicolumn{1}{c|}{61.11$\pm$0.2}       
				& \multicolumn{1}{c}{84.36$\pm$0.2}            & \multicolumn{1}{c}{82.52$\pm$0.3}             & \multicolumn{1}{c}{64.19$\pm$0.3}               & \multicolumn{1}{c}{67.39$\pm$0.1}            & \multicolumn{1}{c}{74.29$\pm$0.2}\\
        \quad 	FedAvg\cite{Mcmahan2017}         & \multicolumn{1}{c}{\underline{48.36}$\pm$0.2}          & \multicolumn{1}{c}{\underline{42.64}$\pm$0.5}              & \multicolumn{1}{c}{\underline{48.84}$\pm$0.1}              & \multicolumn{1}{c}{\underline{49.49}$\pm$0.5}              & 
				\multicolumn{1}{c|}{\underline{49.72}$\pm$0.5}            & \multicolumn{1}{c}{\underline{74.16}$\pm$0.2}          & \multicolumn{1}{c}{68.25$\pm$0.1}                 & \multicolumn{1}{c}{\underline{77.75}$\pm$0.2}              & \multicolumn{1}{c}{\underline{78.17}$\pm$0.1}        & 
				\multicolumn{1}{c|}{\underline{81.35}$\pm$0.1}      & 
				\multicolumn{1}{c}{\underline{87.64}$\pm$0.2}   & \multicolumn{1}{c}{\underline{85.42}$\pm$0.0}          & \multicolumn{1}{c}{\underline{87.32}$\pm$0.1}                 & \multicolumn{1}{c}{\underline{87.27}$\pm$0.2}              & \multicolumn{1}{c}{\underline{88.25}$\pm$0.2}              \\               

	\quad q-FFL\cite{li2019fair}           & \multicolumn{1}{c}{46.22$\pm$1.5}          & \multicolumn{1}{c}{41.40$\pm$0.2}               & \multicolumn{1}{c}{35.00$\pm$0.9}              & \multicolumn{1}{c}{37.41$\pm$1.1}              & 
				\multicolumn{1}{c|}{37.81$\pm$0.4}   & \multicolumn{1}{c}{69.61$\pm$0.6}          & \multicolumn{1}{c}{54.71$\pm$1.52}              & \multicolumn{1}{c}{34.15$\pm$0.6}              & \multicolumn{1}{c}{44.69$\pm$1.3}              & 
				\multicolumn{1}{c|}{56.92$\pm$1.1}   & \multicolumn{1}{c}{85.44$\pm$0.2}          & \multicolumn{1}{c}{82.93$\pm$0.5}                & \multicolumn{1}{c}{65.51$\pm$2.4}              & \multicolumn{1}{c}{69.61$\pm$2.1}              & 
				\multicolumn{1}{c}{78.96$\pm$0.8}\\

	\quad CFFL\cite{lyu2020collaborative}           & \multicolumn{1}{c}{47.94$\pm$0.6}          & \multicolumn{1}{c}{42.12$\pm$0.3}             & \multicolumn{1}{c}{44.44$\pm$0.9}              & \multicolumn{1}{c}{48.44$\pm$0.3}              & 
				\multicolumn{1}{c|}{47.56$\pm$1.2}   & \multicolumn{1}{c}{72.68$\pm$0.2}          & \multicolumn{1}{c}{\underline{69.66}$\pm$0.7}                 & \multicolumn{1}{c}{75.89$\pm$0.9}              & \multicolumn{1}{c}{75.21$\pm$0.3}              & 
				\multicolumn{1}{c|}{77.01$\pm$0.9}   & \multicolumn{1}{c}{87.40$\pm$0.4}          & \multicolumn{1}{c}{85.29$\pm$0.4}            & \multicolumn{1}{c}{86.16$\pm$1.0}              & \multicolumn{1}{c}{87.23$\pm$0.2}              & 
				\multicolumn{1}{c}{87.11$\pm$0.8}\\
				
	\quad CGSV\cite{xu2021gradient}             & \multicolumn{1}{c}{34.60$\pm$0.7}          & \multicolumn{1}{c}{39.33$\pm$0.5}               & \multicolumn{1}{c}{47.22$\pm$0.6}              & \multicolumn{1}{c}{44.12$\pm$0.5}              & 
				\multicolumn{1}{c|}{39.31$\pm$0.7}   & \multicolumn{1}{c}{65.92$\pm$0.6}          & \multicolumn{1}{c}{65.96$\pm$0.3}            & \multicolumn{1}{c}{73.00$\pm$1.9}              & \multicolumn{1}{c}{75.16$\pm$0.2}      & \multicolumn{1}{c|}{77.42$\pm$0.5}         & \multicolumn{1}{c}{83.16$\pm$0.3}          & \multicolumn{1}{c}{82.41$\pm$1.3}                 & \multicolumn{1}{c}{85.03$\pm$2.9}              & \multicolumn{1}{c}{84.73$\pm$3.2}              & 
				\multicolumn{1}{c}{76.51$\pm$4.2}\\

	\quad {FedAVE\cite{wang2024fedave}}             & \multicolumn{1}{c}{46.51$\pm$0.2}          & \multicolumn{1}{c}{35.18$\pm$1.5}                & \multicolumn{1}{c}{46.60$\pm$0.6}              & \multicolumn{1}{c}{39.20$\pm$1.2}              & 
				\multicolumn{1}{c|}{40.60$\pm$1.8}   & \multicolumn{1}{c}{70.80$\pm$0.5}       &\multicolumn{1}{c}{65.10$\pm$0.6}        &\multicolumn{1}{c}{73.44$\pm$0.6}              & \multicolumn{1}{c}{73.43$\pm$0.3}              & 
				\multicolumn{1}{c|}{75.48$\pm$0.7}   & \multicolumn{1}{c}{86.18$\pm$0.6}          & \multicolumn{1}{c}{79.86$\pm$1.2}               & \multicolumn{1}{c}{76.66$\pm$0.8}              & \multicolumn{1}{c}{80.90$\pm$1.3}              & 
				\multicolumn{1}{c}{67.74$\pm$0.8} \\

	\quad Ours      & \multicolumn{1}{c}{\textbf{48.61}$\pm$0.2}          & \multicolumn{1}{c}{\textbf{44.16}$\pm$0.2}              & \multicolumn{1}{c}{\textbf{49.06}$\pm$0.6}              & \multicolumn{1}{c}{\textbf{50.01}$\pm$0.2}              & 
				\multicolumn{1}{c|}{\textbf{49.85}$\pm$0.3}  & 
				\multicolumn{1}{c}{\textbf{74.84}$\pm$0.2}    & \multicolumn{1}{c}{\textbf{70.51}$\pm$0.8}       & \multicolumn{1}{c}{\textbf{78.06}$\pm$0.1}           & \multicolumn{1}{c}{\textbf{78.55}$\pm$0.1}              & \multicolumn{1}{c|}{\textbf{81.95}$\pm$0.4}              & 
				\multicolumn{1}{c}{\textbf{87.88}$\pm$0.2}   & \multicolumn{1}{c}{\textbf{85.61}$\pm$0.3}          & \multicolumn{1}{c}{\textbf{87.85}$\pm$0.1}             & \multicolumn{1}{c}{\textbf{87.54}$\pm$0.4}              & \multicolumn{1}{c}{\textbf{88.38}$\pm$0.4}              \\
            \bottomrule
        \end{tabular}
    }
    \caption{~\textbf{Comparison results of the maximum test accuracy ($\%$)} with state-of-the-art methods on three datasets. The reported results are averaged over 5 runs with different random seeds. (A higher value indicates better accuracy.)}
    \label{table2}
\end{table*}

\setlength\parindent{1.5em}\indent \noindent\textbf{Baselines.} We compare FedSAC with the following methods:
(1) 
FedAvg~\cite{Mcmahan2017} distributes the same model to all clients in each FL iteration. In this case, Pearson Correlation Coefficient $\rho()$ in Section~\ref{sec3.1} is uncomputable. To address this and create a personalized model for each client, we follow CFFL~\cite{lyu2020collaborative} and CGSV~\cite{xu2021gradient}, which enables clients to train for an additional epoch at the end of FL algorithm.
(2) q-FFL\cite{li2019fair} enables the reweighting of loss across different clients by the q-parameterized weights, thus reducing the variance in the accuracy distribution and achieving a fairer distribution of accuracy. 
(3) CFFL~\cite{lyu2020collaborative} allocates more gradients to higher reputation client, and the reputation is calculated by the local accuracy and data sizes (or label diversity).
(4) CGSV~\cite{xu2021gradient} assigns more gradients to clients whose local model gradients is more similar to the global gradients.
(5) FedAVE~\cite{wang2024fedave} assigns more gradients to clients whose data distribution information is more similar to the ideal dataset.
(6) Standalone~\cite{lyu2020collaborative} trains local models alone without collaboration. Particularly, to evaluate more fairly, we make all algorithms distribute rewards based on client contributions rather than the calculated reputations.

\begin{figure*}[!ht]
	\centering
	\includegraphics[width=1.0\textwidth]{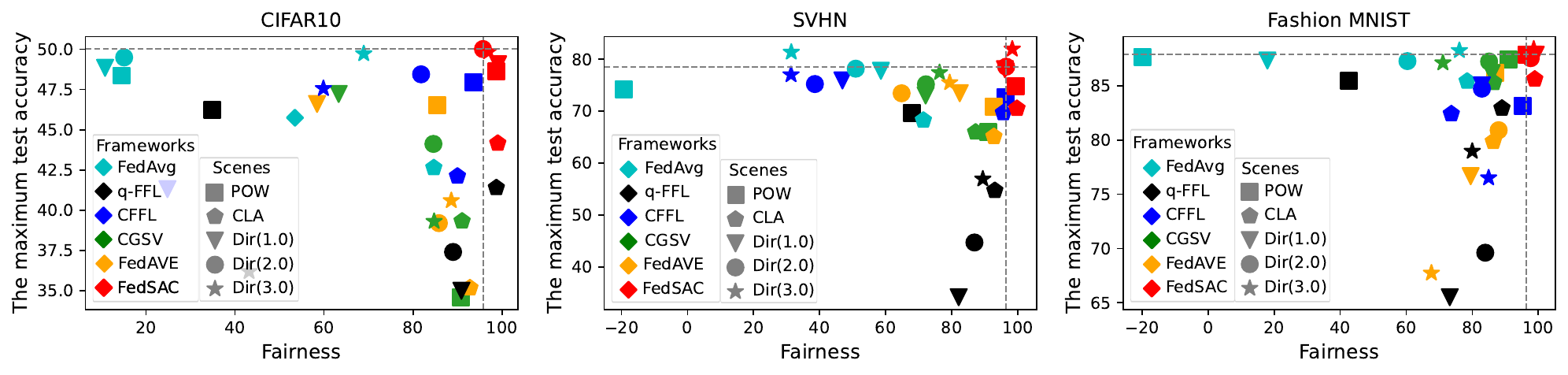}
	\caption{~\textbf{Comparison results of overall performance to achieve bounded collaborative fairness} with state-of-the-art methods in CIFAR10 (left), SVHN (middle), and Fashion MNIST (right). (The closer the point is to the upper-right corner, the better the performance.)}
	\label{fig.3}
\end{figure*}

\setlength\parindent{1.5em}\indent \noindent\textbf{Hyper-Parameters.} We tune all hyper-parameters in datasets by using grid search with FedAvg~\cite{Mcmahan2017} and subsequently apply the optimal parameters obtained from the validation dataset. The batch size is $B$ = 64 for SVHN and $B$ = 32 for both FashionMNIST and CIFAR10. The optimal parameters for SVHN, Cifar10, and FashionMNIST of six scenarios are $E$ = \{15, 20\}, $\eta$ = \{0.05, 0.1\}, $\beta$ = \{3, 5, 10\}; $E$ = \{15, 20\}, $\eta$ = \{0.03, 0.05\}, $\beta$ = \{3, 5, 10\}, and $E$ = 20, $\eta$ = \{0.03, 0.05\}, $\beta$ = \{1, 10, 20, 25\}, respectively. For comparison, we select the best fairness achieved by each method. The effects of hyper-parameter $\beta$ on FedSAC are detailed in Appendix D (the smaller $\beta$ is, the higher the accuracy achieved by FedSAC). More details about the hyper-parameters are put in Appendix F.\par

\setlength\parindent{1.5em}\textbf{Implementation.} All experiments are run on a 64 GB-RAM Ubuntu 18.04.6 server with Intel(R) Xeon(R) CPU E5-2630 v4 @ 2.20GHz and 1 NVidia(R) 2080Ti GPUs. 

\subsection{Experimental Results}\label{sec.3.1}

\setlength{\parindent}{0pt}\textbf{Fairness (RQ1).} To evaluate the FedSAC fairness, we compared it with a few baselines on three datasets. Table \ref{tabel1} shows the fairness metrics according to $Definition$ \ref{definition1}. Standalone~\cite{lyu2020collaborative} trains local models alone without collaboration, which represents the clients' contributions. Table \ref{tabel1} indicates that our proposed dynamic submodel allocation mechanism achieves a fairness score above 95.73\% on all datasets, while the FedAvg performs poorly with the lowest fairness score of -19.83\%. On three datasets, the fairness of algorithms (i.e., CFFL, CGSV, and FedAVE) exceeds 73.65\% for the POW and CLA scene. In these scenarios, the client’s contribution varies greatly and is mainly related to the amount of data or diverse labels of data. For the DIR scene, the data distribution among clients is significantly uneven, resulting in a high degree of non-iid settings and clients with relatively similar contributions. Consequently, CFFL, CGSV, and FedAVE show low fairness, as the rewards received by clients tend to be indistinguishable. In particular, in DIR (1.0) of CIFAR10, our method outperforms CFFL, CGSV and FedAVE by 69.24\%, 35.85\%, and 40.71\%, respectively.\par
\setlength\parindent{1.5em}Table \ref{tabel1} demonstrates that the proposed FedSAC outperforms the state-of-the-art approaches in fairness, and validated the effectiveness of our method: high-contributing clients obtain high-performance models. Figure~\ref{fig.3} shows the comparison results of overall performance to achieve bounded collaborative fairness with state-of-the-art methods in CIFAR10 (left), SVHN (middle), and Fashion MNIST (right). 
\textbf{Obviously}, FedSAC outperforms all baselines in terms of fairness.

\begin{table*}
    \centering
    \resizebox{1.0\linewidth}{!}{
        \small
        \begin{tabular}{@{}l|c|ccccc|ccccc|ccccc@{}}
            \toprule
            \quad  Dataset        & \multicolumn{5}{c|}{CIFAR10} &\multicolumn{5}{c|}{SVHN} &\multicolumn{5}{c}{Fashion MNIST}\\ 
             \midrule
		\quad Scene & \multicolumn{1}{c}{POW}            & \multicolumn{1}{c}{CLA}                & \multicolumn{1}{c}{DIR(1.0)}               & \multicolumn{1}{c}{DIR(2.0)}            & \multicolumn{1}{c|}{DIR(3.0)}     
				& \multicolumn{1}{c}{POW}            & \multicolumn{1}{c}{CLA}                  & \multicolumn{1}{c}{DIR(1.0)}               & \multicolumn{1}{c}{DIR(2.0)}            & \multicolumn{1}{c|}{DIR(3.0)}& \multicolumn{1}{c}{POW}            & \multicolumn{1}{c}{CLA}                  & \multicolumn{1}{c}{DIR(1.0)}               & \multicolumn{1}{c}{DIR(2.0)}            & \multicolumn{1}{c}{DIR(3.0)}\\

            \midrule
             
            \quad 	$w/o$ $allocation$        & \multicolumn{1}{c}{79.99}          & \multicolumn{1}{c}{96.47}                & \multicolumn{1}{c}{10.93}              & \multicolumn{1}{c}{43.81}              & 
				\multicolumn{1}{c|}{-14.61}            & \multicolumn{1}{c}{96.88}          & \multicolumn{1}{c}{94.88}                & \multicolumn{1}{c}{31.86}              & \multicolumn{1}{c}{64.94}              & 
				\multicolumn{1}{c|}{82.58}  & \multicolumn{1}{c}{91.05}          & \multicolumn{1}{c}{96.09}                & \multicolumn{1}{c}{84.57}              & \multicolumn{1}{c}{79.14}              & 
				\multicolumn{1}{c}{65.53}\\
    
            \quad $w/o$ $aggregation$         & \multicolumn{1}{c}{98.64}          & \multicolumn{1}{c}{98.13}              & \multicolumn{1}{c}{99.22}              & \multicolumn{1}{c}{84.81}              & 
				\multicolumn{1}{c|}{83.59}            & \multicolumn{1}{c}{99.00}          & \multicolumn{1}{c}{99.20}            & \multicolumn{1}{c}{95.66}              & \multicolumn{1}{c}{96.05}              & 
				\multicolumn{1}{c|}{87.65}  & \multicolumn{1}{c}{93.72}          & \multicolumn{1}{c}{97.45}                & \multicolumn{1}{c}{97.32}              & \multicolumn{1}{c}{27.40}              & 
				\multicolumn{1}{c}{91.43}\\
            \quad $FedSAC$        & \multicolumn{1}{c}{\textbf{98.80}}          & \multicolumn{1}{c}{\textbf{99.06}}      & \multicolumn{1}{c}{\textbf{99.14}}              & \multicolumn{1}{c}{\textbf{95.73}}              & 
				\multicolumn{1}{c|}{\textbf{97.01}}   & \multicolumn{1}{c}{\textbf{99.44}}          & \multicolumn{1}{c}{\textbf{99.74}}            & \multicolumn{1}{c}{\textbf{96.09}}              & \multicolumn{1}{c}{\textbf{96.48}}              & 
				\multicolumn{1}{c|}{\textbf{98.32}}    & \multicolumn{1}{c}{\textbf{96.35}}          & \multicolumn{1}{c}{\textbf{98.93}}                & \multicolumn{1}{c}{\textbf{99.23}}              & \multicolumn{1}{c}{\textbf{97.71}}              & 
				\multicolumn{1}{c}{\textbf{98.62}} \\

            \bottomrule
        \end{tabular}
    }
    \caption{~\textbf{Ablation studies on FedSAC for fairness} $\rho\in[-100, 100]$ on three public benchmarks. A higher $\rho$ denotes better fairness.}
    \label{table4}
\end{table*}

\begin{table*}
    \centering
    \resizebox{1.0\linewidth}{!}{
        \small
        \begin{tabular}{@{}l|c|ccccc|ccccc|ccccc@{}}
            \toprule
            \quad  Dataset        & \multicolumn{5}{c|}{CIFAR10} &\multicolumn{5}{c|}{SVHN} &\multicolumn{5}{c}{Fashion MNIST}\\ 
             \midrule
		\quad Scene & \multicolumn{1}{c}{POW}            & \multicolumn{1}{c}{CLA}                & \multicolumn{1}{c}{DIR(1.0)}               & \multicolumn{1}{c}{DIR(2.0)}            & \multicolumn{1}{c|}{DIR(3.0)}     
				& \multicolumn{1}{c}{POW}            & \multicolumn{1}{c}{CLA}                  & \multicolumn{1}{c}{DIR(1.0)}               & \multicolumn{1}{c}{DIR(2.0)}            & \multicolumn{1}{c|}{DIR(3.0)}& \multicolumn{1}{c}{POW}            & \multicolumn{1}{c}{CLA}                  & \multicolumn{1}{c}{DIR(1.0)}               & \multicolumn{1}{c}{DIR(2.0)}            & \multicolumn{1}{c}{DIR(3.0)}\\

            \midrule
             
            \quad 	$w/o$ $allocation$        & \multicolumn{1}{c}{47.68}          & \multicolumn{1}{c}{42.46}        & \multicolumn{1}{c}{40.02}              & \multicolumn{1}{c}{47.00}              & \multicolumn{1}{c|}{44.09}            & \multicolumn{1}{c}{69.98}          & \multicolumn{1}{c}{66.95}                  & \multicolumn{1}{c}{66.25}              & \multicolumn{1}{c}{68.14}              & 
				\multicolumn{1}{c|}{74.02}  & \multicolumn{1}{c}{87.48}          & \multicolumn{1}{c}{85.00}                  & \multicolumn{1}{c}{70.99}              & \multicolumn{1}{c}{73.13}              & 
				\multicolumn{1}{c}{83.43}\\
    
            \quad $w/o$ $aggregation$        & \multicolumn{1}{c}{48.03}          & \multicolumn{1}{c}{43.62}              & \multicolumn{1}{c}{48.23}              & \multicolumn{1}{c}{49.46}              & \multicolumn{1}{c|}{48.61}            & \multicolumn{1}{c}{73.81}          & \multicolumn{1}{c}{68.91}                & \multicolumn{1}{c}{77.35}              & \multicolumn{1}{c}{72.89}              & 
				\multicolumn{1}{c|}{80.46}   & \multicolumn{1}{c}{87.74}          & \multicolumn{1}{c}{85.95}                  & \multicolumn{1}{c}{86.90}              & \multicolumn{1}{c}{85.72}              & 
				\multicolumn{1}{c}{86.74}\\
            \quad $FedSAC$           & \multicolumn{1}{c}{\textbf{48.61}}          & \multicolumn{1}{c}{\textbf{44.16}}                  & \multicolumn{1}{c}{\textbf{49.06}}              & \multicolumn{1}{c}{\textbf{50.01}}              & 
				\multicolumn{1}{c|}{\textbf{49.85}}   & \multicolumn{1}{c}{\textbf{74.84}}          & \multicolumn{1}{c}{\textbf{70.51}}            & \multicolumn{1}{c}{\textbf{78.06}}              & \multicolumn{1}{c}{\textbf{78.55}}              & 
				\multicolumn{1}{c|}{\textbf{81.95}}   & \multicolumn{1}{c}{\textbf{87.88}}          & \multicolumn{1}{c}{\textbf{85.61}}                  & \multicolumn{1}{c}{\textbf{87.85}}              & \multicolumn{1}{c}{\textbf{87.54}}              & 
				\multicolumn{1}{c}{\textbf{88.38}} \\

            \bottomrule
        \end{tabular}
    }
    \caption{~\textbf{Ablation studies on FedSAC for the maximum test accuracy (\%)} on three public benchmarks.}
    \label{table5}
\end{table*}

\begin{figure}[!ht]
	\centering
	\includegraphics[width=0.46\textwidth]{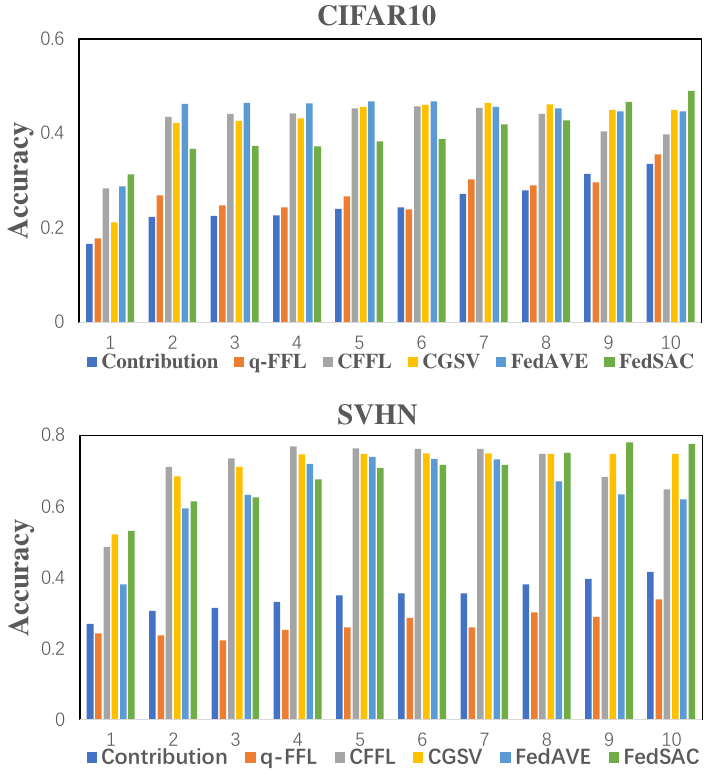}
	\caption{~\textbf{Comparison results of test accuracy using the data partition of DIR (1.0)} with state-of-the-art methods in CIFAR10 (up) and SVHN (down). Results of other scenes are in Appendix D.} 
	\label{fig.4}
\end{figure}

\setlength{\parindent}{1.5em}\textbf{Predictive performance (RQ2).} To effectively assess the predictive performance of algorithms, we present our highest test accuracies in comparison with all baseline methods in Table \ref{table2}. These results demonstrate the ability of the algorithms to reward high-contributing clients with high-performance.
First, comparing the accuracy of FedSAC with Standalone (i.e., contribution) reveals that FedSAC significantly outperforms Standalone.
Second, among the POW scene, FedSAC achieves the highest performance in CIFAR10, SVHN and Fashion MNIST with accuracies of 48.61\%, 74.84\%, and 87.88\%, respectively. 
Third, for the CLA scene on three datasets, the highest accuracy is obtained by FedSAC, surpassing FedAvg by at least 0.19\%. In addition, in the extremely non-iid setting (e.g., DIR (1.0) of SVHN), our method outperforms CFFL, CGSV, and FedAVE by 2.17\%, 5.06\%, and 4.62\%, respectively.
Finally, for the DIR(2.0), and DIR (3.0) scenes, FedSAC achieves comparable performance to baseline methods in terms of accuracy. Specially, the notably poor accuracy of q-FFL appears attributed to its mechanism that offers the same reward to all clients, without adapting these rewards based on individual client contributions.  Figure~\ref{fig.4} illustrates the distributions of contributions and allocated rewards under scenes (i.e., DIR (1.0)) in CIFAR10 comparing FedSAC against baseline methods. It demonstrates that FedSAC not only guarantees BCF but also enables clients to receive rewards that exceed their contributions (i.e., Standalone). \textbf{In short}, FedSAC outperforms all baselines in terms of accuracy. More results under different scenarios (i.e., CLA scene) on CIFAR10 and SVHN are presented in Appendix B. \par

Figure \ref{figure 5.} illustrates the changes in clients' test accuracy as the number of communication rounds increases in the POW, and CLA data partition of CIFAR10 and SVHN. Owing to the varying data sizes and diversity of labels owned by clients in FL, their contributions to the system exhibit significant differences. As shown in Figure \ref{figure 5.}, our proposed FedSAC, underpinned by a theoretical guarantee, aims to reward high-contributing clients with high-performance submodels by maintaining consistency in local models. As a result, each client will converge to a different model and achieve varying levels of performance.
\begin{figure}[!ht]
	\centering
	\renewcommand\thefigure{5}
	\includegraphics[scale=0.36]{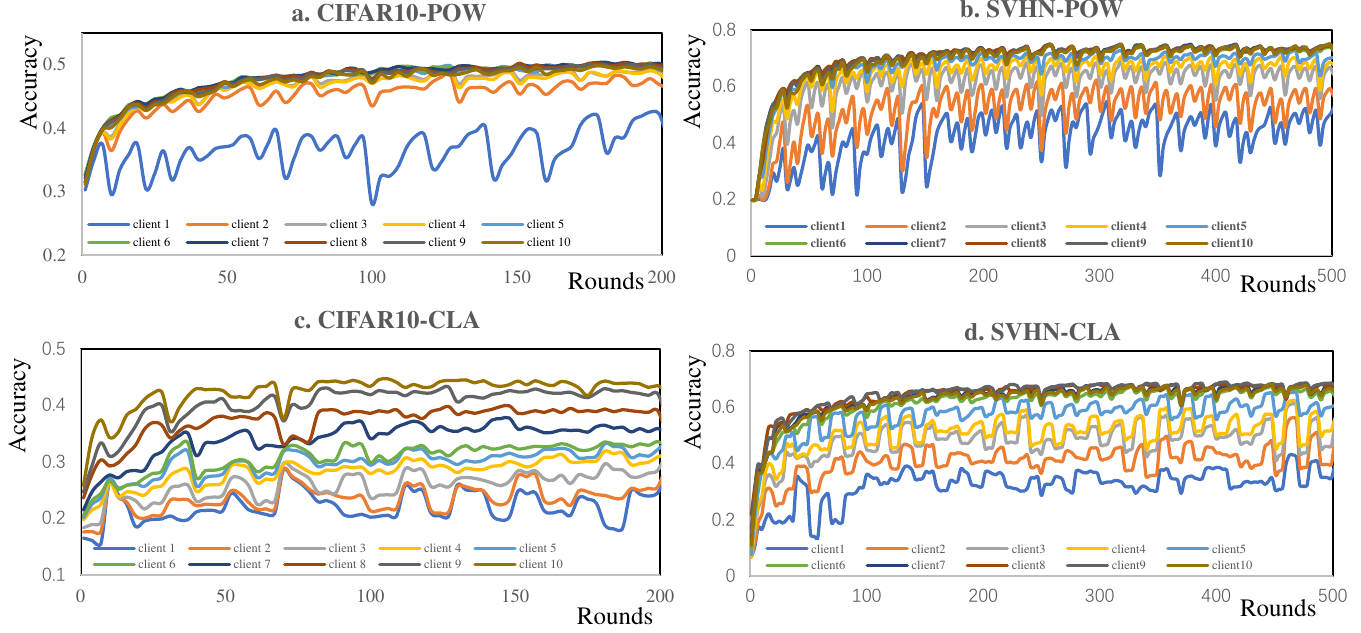}
	\caption{The test accuracy achieved by clients during training for CIFAR10 (left) and SVHN (right) in each round, under the setting of POW and CLA.}
	\label{figure 5.}
\end{figure}


\indent \noindent\textbf{Ablation study (RQ3).} To evaluate the effectiveness of two proposed modules in FedSAC, a series of ablation experiments are carried out on three public benchmarks with 10 clients, as shown in Table \ref{table4} and \ref{table5}. The operation of eliminating neuron importance, denoted as $w/o$ $allocation$, aims to treat all neurons equally and allocate submodels based on their contributions. $w/o$ $aggregation$ denotes removing the dynamic aggregation module, which uses the traditional FedAvg aggregation method to train. The effectiveness of the submodel allocation module is demonstrated in Table \ref{table4}, indicating that this module can reward high-contribution clients to obtain high-performance models. In particular, our method has significantly improved the fairness measure by 88.21\% on the DIR (1.0) scene of the CIFAR10 dataset. Table \ref{table5} shows the results of the proposed dynamic aggregation module, which implies that this module can effectively aggregate submodels with different sizes, thereby further improving the overall performance of the local models. \textbf{Thus}, the ablation study demonstrates that the two designed modules in FedSAC are crucial and significant in enhancing bounded collaborative fairness.


\setlength{\parindent}{2em}\textbf{In summary}, all experimental results show that both fairness and model accuracy are of significance for bounded collaborative fairness, and our FedSAC outperforms all baseline methods in both fairness and model accuracy.

\section{Conclusion}

\setlength{\parindent}{0pt}In this work, we introduce a novel FL framework named FedSAC that allocates submodels based on their contributions, thereby ensuring bounded collaborative fairness and attaining superior local accuracy while maintaining the consistency in local models. Our method ensures that  high-contributing clients can be rewarded with high-performance submodels, which in turn enhances the overall model accuracy. The experiments on three datasets show that FedSAC exhibits a distinct advantage over baseline methods in terms of fairness and accuracy.
In the future, we aim to investigate the implementation of FedSAC on large models.

\bibliographystyle{ACM-Reference-Format}
\bibliography{sample-base}

\appendix

\clearpage

\section*{A. Proof of Theorem 2}
Let $I_E$ be the set of global synchronization steps, i.e., $I_E=\{nE|n=1, 2, ...\}$. For convenience, We define $v_i^{t+1}$ as the immediate result of one step SGD update from $\theta_i^t$, i.e., $v_i^{t+1} = \theta_i^t - \eta_t\nabla F_i(\theta_i^t, \xi_i^t)$. $\bar{g}_t = \sum_{i=1}^{N}\frac{\nabla F_i(\theta_i^t)}{p_i}$ and $g_t = \sum_{i=1}^{N}\frac{\nabla F_i(\theta_i^t, \xi_i^t)}{p_i}$. Therefore, $\bar{v}_{t+1} = \bar{\theta}_t - \eta_t g_t$ and $Eg_t = \bar{g}_t$.

\setlength{\parindent}{0pt}
\begin{lemma}
\setcounter{lemma}{0}
\label{lamma1}
(Result of one step SGD). Assume ASSUMPTION~\ref{Assumption1} and ASSUMPTION~\ref{Assumption2}. If $\eta_t \leqslant \frac{1}{4L}$, we have
\begin{equation}
\label{eq21}
\begin{split}
E\|\bar{v}_{t+1} - \theta^* \|^2 &\leqslant (1-\eta_t \mu)E\|\bar{\theta}_t -\theta^*\|^2 + \eta_t^2E\| g_t - \bar{g}_t \|^2 \\
&\quad+ 6L\eta_t^2 \Gamma + 2E\sum_{i=1}^N\frac{\|\bar{\theta}_t - \theta_i^t \|^2}{p_i},
\end{split}
\end{equation}
\end{lemma}
where $\Gamma = F^* - \sum_{i=1}^N\frac{F_i^*}{p_i}$. LAMMA~\ref{lamma1} has been made by~\cite{li2019convergence}.

\setlength{\parindent}{0.5cm}For Assumption~\ref{Assumption3}, the variance of stochastic gradients within device $i$ is constrained by $\sigma_i^2$. Consequently,

\begin{equation}
\label{eq22}
\begin{split}
E\|g_t - \bar{g}_t \|^2 &= E\|\sum_{i=1}^N\frac{1}{p_i}(\nabla F_k(\theta_i^t, \xi_i^t) - \nabla F_i(\theta_i^t))\|^2 \\
&=\sum_{i=1}^N\frac{1}{p_i^2} E\|\nabla F_i(\theta_i^t, \xi_i^t) - \nabla F_i(\theta_i^t)\|^2 \\
&\leqslant\sum_{i=1}^N\frac{1}{p_i^2}\sigma_i^2.
\end{split}
\end{equation}

As FedSAC requires communication each $E$ steps. We let $\eta_t \leqslant 2\eta_{t+E}$. Therefore, for any $t\ge0$, there exists a $t_0 \leqslant t$, such that $t - t_0 \leqslant E-1$ and $\theta_i^{t_0} = \bar{\theta}_{t_0}$ for all $k = 1, 2, ..., N$. Then 

\begin{equation}
\label{eq23}
\begin{split}
E\sum_{i=1}^N \frac{1}{p_i}\|\bar{\theta}_t - \theta_i^t \|^2 &=E\sum_{i=1}^N\frac{1}{p_i}\|(\theta_i^t - \bar{\theta}_{t_0}) - (\bar{\theta}_t - \bar{\theta}_{t_0})\|^2\\
&\leqslant E \sum_{i=1}^N \frac{1}{p_i}\|\theta_i^t - \bar{\theta}_{t_0}\|^2 \\
&\leqslant E\sum_{t=t_0}^{t-1}(E-1)\eta_t^2 \|\nabla F_k(\theta_i^t, \xi_i^t)\|^2 \\
&\leqslant \sum_{t=t_0}^{t-1}(E-1)\eta_{t_0}^2 G^2\\
&\leqslant \eta_{t_0}^2(E-1)^2 G^2 \\
&\leqslant 4\eta_t^2(E-1)^2 G^2.
\end{split}
\end{equation}
Here in lines 1252-1256, we use $E\|X-EX\|^2 \leqslant E\|X\|^2$ where $X=\theta_i^t-\bar{\theta}_{t_0}$ with probability $\frac{1}{p_i}$. In the lines 1256-1259, we use Jensen inequality:
\begin{equation}
\begin{split}
\|\theta_i^t - \bar{\theta}_{t_0}\| &= \|\sum_{t=t_0}^{t-1} \eta_t \nabla F_i(\theta_i^t, \xi_i^t)\|^2 \\
&\leqslant (t-t_0)\sum_{t-t_0}\sum_{t-t_0}^{t-1}\eta_t^2\|\nabla F_i(\theta_i^t, \xi_i^t)\|^2.
\end{split}
\end{equation}
In lines 1259-1262, we utilize $\eta_t \leqslant \eta_{t_0}$ for $t\ge t_0$ and $E\| \nabla F_k(\theta_i^t, \xi_i^t)\|^2 \leqslant G^2$ for $i = 1, 2, ..., N$. In the lines 1263-1265, we use $\eta_{t_0} \leqslant 2\eta_{t_0+E} \leqslant 2\eta_t$ for $t_0 \leqslant t \leqslant t_0 +E$.

Let $\triangle_t = E\|\bar{\theta}_t - \theta^*\|$. Fromr Eq.(\ref{eq21}), Eq. (\ref{eq22}), and Eq. (\ref{eq23}), it follows that
\begin{equation}
\begin{split}
\triangle_{t+1} &\leqslant (1-\eta_t\mu)\triangle_t + \eta_t^2 \sum_{i=1}^N \frac{\sigma^2}{p_i^2} + 6L\eta_t^2\Gamma + 8\eta_t^2(E-1)^2G^2\\
&\leqslant (1-\eta_t\mu)\triangle_t + \eta_t^2 \underbrace{(\sum_{i=1}^N \frac{\sigma^2}{p_i^2} + 6L\Gamma + 8(E-1)^2G^2)}_{\text{$B$}}
\end{split}
\end{equation}

For a diminishing stepsize, $\eta_t = \frac{\kappa}{t+\gamma}$ for some $\kappa > \frac{1}{\mu}$ and $\gamma > 0$ such that $\eta_1 \leqslant min\{ \frac{1}{\mu}, \frac{1}{4L}\} = \frac{1}{4L}$ and $\eta_t\leqslant 2\eta_{t+E}$. We will prove $\triangle \leqslant \frac{v}{\gamma+t}$ by induction, where $v = max \{ \frac{\kappa^2B}{\kappa\mu-1}, (\gamma+1)\triangle_1\}$. Firstly, the definition of $v$ guarantees its applicability for $t=1$. Assuming the conclusion holds for some $t$, it follows that 
\begin{equation}
\begin{split}
\triangle_{t+1} &\leqslant (1-\eta_t\mu)\triangle_t + \eta_t^2 B\\
&\leqslant(1-\frac{\kappa\mu}{t+\gamma})\frac{v}{t+\gamma} + \frac{\kappa^2B}{(t+\gamma)^2}\\
&=\frac{t+\gamma-1}{(t+\gamma)^2}v + [\frac{\kappa^2B}{(t+\gamma)^2} - \frac{\kappa\mu-1}{(t+\gamma)^2}v]\\
&\leqslant\frac{t+\gamma-1}{(t+\gamma)^2}v  + \frac{\kappa^2B}{(t+\gamma)^2} - \frac{\kappa^2B}{(t+\gamma)^2} \underbrace{-\frac{\kappa\mu-1}{(t+\gamma)^2(\gamma+1)\triangle_1}}_{\leqslant0}\\
&\leqslant \frac{v}{t+\gamma-1}
\end{split}
\end{equation}

\setlength{\parindent}{0cm}Then by the $L$-smoothness (ASSUMPTION~\ref{Assumption1}) of $F$,
\begin{equation}
\label{eq27}
\begin{split}
E[F(\bar{\theta}_T)] - F^* &\leqslant (\bar{\theta}_T - \theta^*)^T \underbrace{\nabla F_i(\theta^*)}_{=0} + \frac{L}{2}\|\bar{\theta}_T - \theta^*\|_2^2\\
&= \frac{L}{2}\triangle_T\\
&\leqslant \frac{L}{2}\frac{v}{\gamma+T}
\end{split}
\end{equation}

We let $\kappa=\frac{2}{\mu}$, $\gamma = max\{8\frac{L}{\mu}, E\} -1$. In the lines 1293, we have
\begin{equation}
\label{eq28}
\begin{split}
v&=max \{ \frac{\kappa^2B}{\kappa\mu-1}, (\gamma+1)\triangle_1\}\\
&\leqslant \frac{\kappa^2B}{\kappa\mu-1} + (\gamma+1)\triangle_1\\
&\leqslant \frac{4B}{\mu^2} + (\gamma + 1)\triangle_1
\end{split}
\end{equation}

Substituting Eq.~\ref{eq28} into Eq.~\ref{eq27}, we obtain
\begin{equation}
\begin{split}
0\leqslant \lim_{T\to \infty} E[F(\bar{\theta}_T)] - F^* \leqslant \lim_{T \to \infty}[\frac{L}{\gamma+T}(\frac{2B}{\mu^2}+ \frac{\gamma+1}{2}\triangle_1)] = 0
\end{split}
\end{equation}

Therefore, $\lim\limits_{T\to \infty} E[F(\bar{\theta}_T)] - F^* = 0$.


\begin{figure}[!ht]
	\centering
	\renewcommand\thefigure{6}
	\includegraphics[scale=0.33]{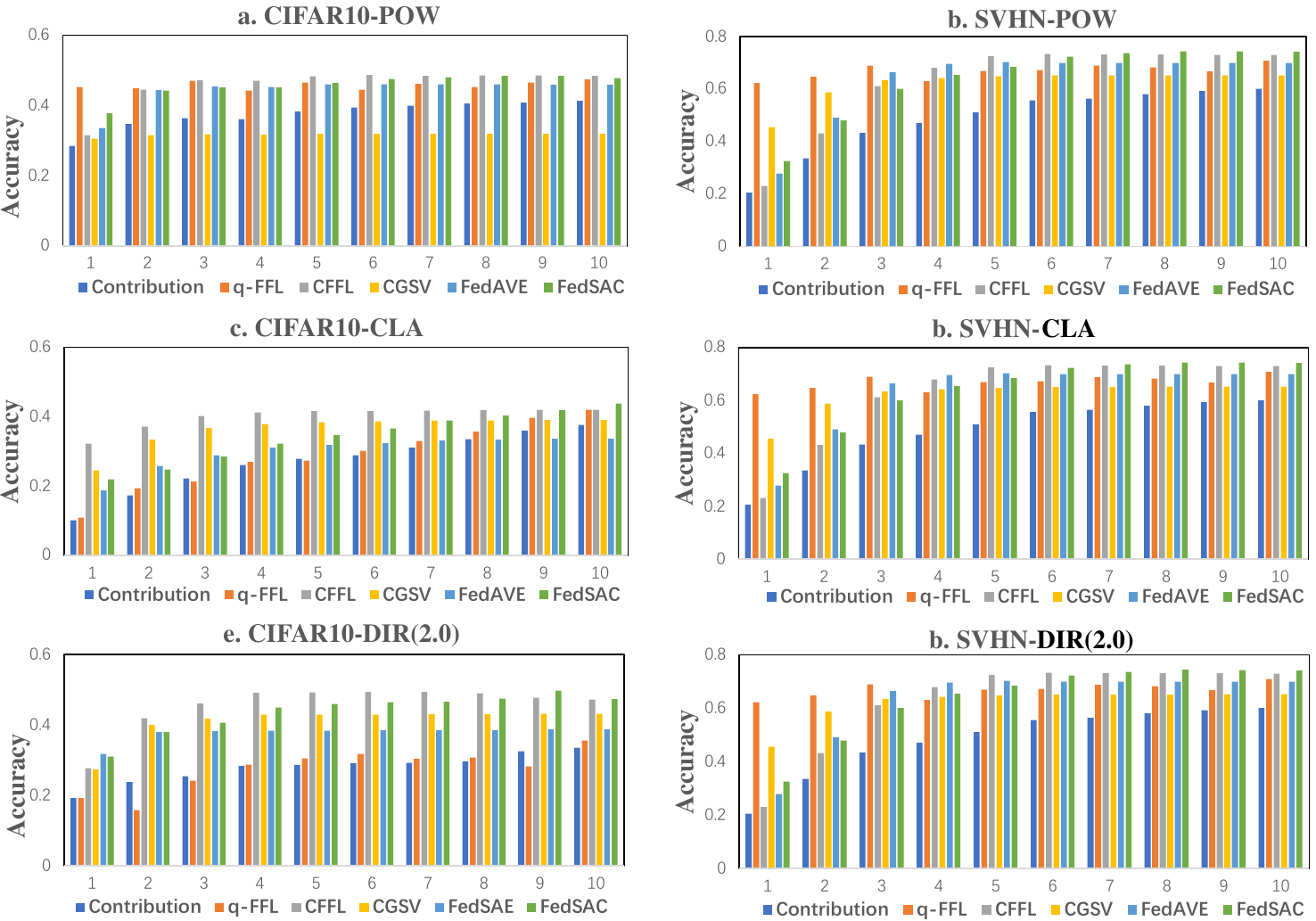}
	\caption{~\textbf{Comparison results of test accuracy using the scene of POW, CLA, and DIR(2.0) with state-of-the-art methods in CIFAR10 (left) and SVHN (right). In terms of accuracy across clients, FedSAC exhibits the highest level of consistency with the contribution.}}
	\label{figure 6.}
\end{figure} 


\begin{table*}[!ht]
	\centering
	\resizebox{1.0\linewidth}{!}{
      \renewcommand\arraystretch{1.3}
		\large
		\begin{tabular}{@{}l|c|ccccccccc|ccccc@{}}
			\toprule
			\quad  Dataset        & \multicolumn{9}{c|}{CIFAR10}& \multicolumn{5}{c}{Fashion MNIST}\\ 

			\midrule
			\quad  No. Clients        & \multicolumn{5}{c|}{20} &\multicolumn{2}{c|}{40} &\multicolumn{2}{c|}{60}& \multicolumn{5}{c}{20}\\ 
   
			\midrule
			\quad Scene & \multicolumn{1}{c}{POW}            & \multicolumn{1}{c}{CLA}                & \multicolumn{1}{c}{DIR(1.0)} & \multicolumn{1}{c}{DIR(2.0)}& \multicolumn{1}{c|}{DIR(3.0)} &  \multicolumn{1}{c}{DIR(1.0)}& \multicolumn{1}{c|}{DIR(2.0)}& \multicolumn{1}{c}{DIR(1.0)}& \multicolumn{1}{c|}{DIR(2.0)}& \multicolumn{1}{c}{POW}&  \multicolumn{1}{c}{CLA}& \multicolumn{1}{c}{DIR(1.0)}& \multicolumn{1}{c}{DIR(2.0)}& \multicolumn{1}{c}{DIR(3.0)}\\

			\midrule
			\quad 	FedAvg\cite{Mcmahan2017}         & \multicolumn{1}{c}{-40.52$\pm$3.3}          & \multicolumn{1}{c}{83.57$\pm$1.4}          & \multicolumn{1}{c}{18.41$\pm$16.3}              & \multicolumn{1}{c}{77.52$\pm$1.5}              & \multicolumn{1}{c|}{44.13$\pm$16.0}                     & \multicolumn{1}{c}{79.53$\pm$13.9}          & \multicolumn{1}{c|}{54.37$\pm$11.1}          & \multicolumn{1}{c}{51.64$\pm$17.8}  & \multicolumn{1}{c|}{55.22$\pm$6.8}   & \multicolumn{1}{c}{-34.64$\pm$5.6} & \multicolumn{1}{c}{84.37$\pm$3.3} & \multicolumn{1}{c}{22.95$\pm$2.6} & \multicolumn{1}{c}{30.39$\pm$6.7} & \multicolumn{1}{c}{56.37$\pm$3.5}    \\               
			
			\quad q-FFL\cite{li2019fair}           & \multicolumn{1}{c}{14.10$\pm$2.8}          & \multicolumn{1}{c}{\underline{98.09}$\pm$0.2}          & \multicolumn{1}{c}{83.98$\pm$2.0}              & \multicolumn{1}{c}{\underline{86.14}$\pm$2.9}         & \multicolumn{1}{c|}{\underline{90.00}$\pm$0.5}      & \multicolumn{1}{c}{\underline{87.81}$\pm$1.6}       & \multicolumn{1}{c|}{80.46$\pm$1.1}                     & \multicolumn{1}{c}{80.65$\pm$2.9}          & \multicolumn{1}{c|}{85.76$\pm$1.2}          & \multicolumn{1}{c}{29.24$\pm$5.3}  & \multicolumn{1}{c}{\underline{98.44}$\pm$2.6}   & \multicolumn{1}{c}{81.75$\pm$4.8} & \multicolumn{1}{c}{77.45$\pm$5.6} & \multicolumn{1}{c}{72.89$\pm$3.9}\\       

   			\quad CFFL\cite{lyu2020collaborative}     & \multicolumn{1}{c}{81.45$\pm$1.0}          & \multicolumn{1}{c}{95.93$\pm$0.8}          & \multicolumn{1}{c}{76.72$\pm$2.1}              & \multicolumn{1}{c}{76.09$\pm$1.3}              & \multicolumn{1}{c|}{63.79$\pm$0.5}                     & \multicolumn{1}{c}{50.33$\pm$1.2}          & \multicolumn{1}{c|}{49.61$\pm$0.6}          & \multicolumn{1}{c}{\underline{86.59}$\pm$1.3}  & \multicolumn{1}{c|}{\underline{87.52}$\pm$0.5}   & \multicolumn{1}{c}{\underline{88.02}$\pm$0.4} & \multicolumn{1}{c}{92.29$\pm$2.5} & \multicolumn{1}{c}{72.74$\pm$2.4} & \multicolumn{1}{c}{78.40$\pm$1.2} & \multicolumn{1}{c}{\underline{75.36}$\pm$1.6}    \\

			\quad CGSV\cite{xu2021gradient}     & \multicolumn{1}{c}{83.30$\pm$1.8}          & \multicolumn{1}{c}{96.80$\pm$0.1}          & \multicolumn{1}{c}{79.85$\pm$0.9}              & \multicolumn{1}{c}{79.73$\pm$1.3}              & \multicolumn{1}{c|}{85.72$\pm$0.4}                     & \multicolumn{1}{c}{82.90$\pm$0.4}          & \multicolumn{1}{c|}{\underline{80.47}$\pm$0.7}          & \multicolumn{1}{c}{85.91$\pm$1.2}  & \multicolumn{1}{c|}{77.91$\pm$0.8}   & \multicolumn{1}{c}{83.85$\pm$0.4} & \multicolumn{1}{c}{94.04$\pm$0.8} & \multicolumn{1}{c}{\underline{88.32}$\pm$2.5} & \multicolumn{1}{c}{\underline{83.68}$\pm$1.1} & \multicolumn{1}{c}{74.15$\pm$1.7}    \\

   			\quad FedAVE\cite{wang2024fedave}     & \multicolumn{1}{c}{\underline{88.46}$\pm$1.5}          & \multicolumn{1}{c}{97.18$\pm$0.6}          & \multicolumn{1}{c}{\underline{87.59}$\pm$1.1}              & \multicolumn{1}{c}{78.43$\pm$0.4}              & \multicolumn{1}{c|}{68.70$\pm$1.3}                     & \multicolumn{1}{c}{45.69$\pm$2.6}          & \multicolumn{1}{c|}{65.60$\pm$1.2}          & \multicolumn{1}{c}{38.92$\pm$1.1}  & \multicolumn{1}{c|}{60.61$\pm$1.3}   & \multicolumn{1}{c}{87.14$\pm$0.6} & \multicolumn{1}{c}{93.97$\pm$1.2} & \multicolumn{1}{c}{78.87$\pm$2.1} & \multicolumn{1}{c}{81.20$\pm$1.5} & \multicolumn{1}{c}{72.73$\pm$0.7}    \\

			\quad Ours   & \multicolumn{1}{c}{\textbf{99.62}$\pm$0.2}          & \multicolumn{1}{c}{\textbf{98.52}$\pm$0.1}          & \multicolumn{1}{c}{\textbf{96.29}$\pm$0.4}              & \multicolumn{1}{c}{\textbf{98.06}$\pm$0.5}              & \multicolumn{1}{c|}{\textbf{95.99}$\pm$0.4}                     & \multicolumn{1}{c}{\textbf{99.37}$\pm$0.5}          & \multicolumn{1}{c|}{\textbf{96.57}$\pm$0.2}          & \multicolumn{1}{c}{\textbf{97.69}$\pm$0.6}  & \multicolumn{1}{c|}{\textbf{95.12}$\pm$0.3}   & \multicolumn{1}{c}{\textbf{97.40}$\pm$0.4} & \multicolumn{1}{c}{\textbf{98.49}$\pm$0.3} & \multicolumn{1}{c}{\textbf{96.52}$\pm$0.8} & \multicolumn{1}{c}{\textbf{97.65}$\pm$0.6} & \multicolumn{1}{c}{\textbf{96.47}$\pm$0.7}    \\       
			\bottomrule
		\end{tabular}
	}
	\renewcommand\thetable{5}
    \caption{~\textbf{Comparison results of fairness $\rho\in[-100, 100]$} with state-of-the-art methods on CIFAR10 and Fashion MNIST. The reported results are averaged over 5 runs with different random seeds. (A higher value indicates better fairness. The best average result is marked in bold. The second-best result is underlined. These notes are the same to others.)}
	\label{tab5}
\end{table*}
\begin{table*}
	\centering
	\resizebox{1.0\linewidth}{!}{
      \renewcommand\arraystretch{1.3}
		\large
		\begin{tabular}{@{}l|c|ccccccccc|ccccc@{}}
			\toprule
			\quad  Dataset        & \multicolumn{9}{c|}{CIFAR10}& \multicolumn{5}{c}{Fashion MNIST}\\ 

			\midrule
			\quad  No. Clients        & \multicolumn{5}{c|}{20} &\multicolumn{2}{c|}{40} &\multicolumn{2}{c|}{60}& \multicolumn{5}{c}{20}\\ 
   
			\midrule
			\quad Scene & \multicolumn{1}{c}{POW}            & \multicolumn{1}{c}{CLA}                & \multicolumn{1}{c}{DIR(1.0)} & \multicolumn{1}{c}{DIR(2.0)}& \multicolumn{1}{c|}{DIR(3.0)} &  \multicolumn{1}{c}{DIR(1.0)}& \multicolumn{1}{c|}{DIR(2.0)}& \multicolumn{1}{c}{DIR(1.0)}& \multicolumn{1}{c|}{DIR(2.0)}& \multicolumn{1}{c}{POW}&  \multicolumn{1}{c}{CLA}& \multicolumn{1}{c}{DIR(1.0)}& \multicolumn{1}{c}{DIR(2.0)}& \multicolumn{1}{c}{DIR(3.0)}\\

			\midrule
			\quad 	Standalone         & \multicolumn{1}{c}{37.15$\pm$0.0}          & \multicolumn{1}{c}{35.25$\pm$0.3}          & \multicolumn{1}{c}{27.82$\pm$0.0}              & \multicolumn{1}{c}{31.83$\pm$0.1}              & \multicolumn{1}{c|}{33.56$\pm$0.1}                     & \multicolumn{1}{c}{32.13$\pm$0.1}          & \multicolumn{1}{c|}{28.86$\pm$0.2}          & \multicolumn{1}{c}{28.58$\pm$0.2}  & \multicolumn{1}{c|}{27.84$\pm$0.1}   & \multicolumn{1}{c}{82.36$\pm$0.1} & \multicolumn{1}{c}{81.17$\pm$0.2} & \multicolumn{1}{c}{66.71$\pm$0.2} & \multicolumn{1}{c}{68.00$\pm$0.2} & \multicolumn{1}{c}{73.63$\pm$0.3}    \\

			\quad 	FedAvg\cite{Mcmahan2017}         & \multicolumn{1}{c}{\underline{46.94}$\pm$0.2}          & \multicolumn{1}{c}{41.14$\pm$0.6}          & \multicolumn{1}{c}{\underline{48.31}$\pm$0.5}              & \multicolumn{1}{c}{\underline{49.00}$\pm$0.5}              & \multicolumn{1}{c|}{\underline{50.13}$\pm$0.3}                     & \multicolumn{1}{c}{\underline{48.19}$\pm$0.3}   & \multicolumn{1}{c|}{\underline{49.57}$\pm$0.3}        & \multicolumn{1}{c}{49.06$\pm$0.1}          & \multicolumn{1}{c|}{\underline{49.30}$\pm$0.3}  & \multicolumn{1}{c}{87.29$\pm$0.3}   & \multicolumn{1}{c}{84.67$\pm$0.2} & \multicolumn{1}{c}{\underline{87.46}$\pm$0.0} & \multicolumn{1}{c}{\underline{88.06}$\pm$0.1} & \multicolumn{1}{c}{88.04$\pm$0.1}    \\               
			
			\quad q-FFL\cite{li2019fair}           & \multicolumn{1}{c}{46.87$\pm$0.2}          & \multicolumn{1}{c}{\underline{41.56}$\pm$0.3}          & \multicolumn{1}{c}{33.77$\pm$0.1}              & \multicolumn{1}{c}{38.17$\pm$0.9}              & \multicolumn{1}{c|}{43.91$\pm$0.6}                     & \multicolumn{1}{c}{38.64$\pm$1.5}          & \multicolumn{1}{c|}{39.37$\pm$0.4}          & \multicolumn{1}{c}{34.01$\pm$0.1}  & \multicolumn{1}{c|}{36.77$\pm$0.5}   & \multicolumn{1}{c}{85.79$\pm$0.2} & \multicolumn{1}{c}{81.10$\pm$0.1} & \multicolumn{1}{c}{68.30$\pm$1.7} & \multicolumn{1}{c}{78.59$\pm$1.8} & \multicolumn{1}{c}{77.69$\pm$1.1}    \\       
			
			\quad CFFL\cite{lyu2020collaborative}     & \multicolumn{1}{c}{46.06$\pm$0.0}          & \multicolumn{1}{c}{39.43$\pm$0.3}          & \multicolumn{1}{c}{45.76$\pm$0.3}              & \multicolumn{1}{c}{48.57$\pm$0.2}              & \multicolumn{1}{c|}{48.42$\pm$0.3}                     & \multicolumn{1}{c}{39.54$\pm$0.3}          & \multicolumn{1}{c|}{39.16$\pm$0.6}          & \multicolumn{1}{c}{42.33$\pm$0.4}  & \multicolumn{1}{c|}{41.06$\pm$0.6}   & \multicolumn{1}{c}{85.88$\pm$1.8} & \multicolumn{1}{c}{81.75$\pm$2.1} & \multicolumn{1}{c}{81.69$\pm$0.6} & \multicolumn{1}{c}{84.16$\pm$0.4} & \multicolumn{1}{c}{87.34$\pm$0.4}    \\       

   			\quad CGSV\cite{xu2021gradient}     & \multicolumn{1}{c}{46.29$\pm$0.2}          & \multicolumn{1}{c}{37.75$\pm$1.4}          & \multicolumn{1}{c}{46.72$\pm$1.0}              & \multicolumn{1}{c}{48.45$\pm$0.3}              & \multicolumn{1}{c|}{49.24$\pm$0.2}                     & \multicolumn{1}{c}{46.75$\pm$0.2}          & \multicolumn{1}{c|}{48.32$\pm$0.1}          & \multicolumn{1}{c}{\underline{49.11}$\pm$0.7}  & \multicolumn{1}{c|}{48.58$\pm$0.3}   & \multicolumn{1}{c}{87.21$\pm$0.2} & \multicolumn{1}{c}{84.25$\pm$0.2} & \multicolumn{1}{c}{85.17$\pm$0.7} & \multicolumn{1}{c}{86.85$\pm$0.3} & \multicolumn{1}{c}{\underline{88.07}$\pm$0.2}    \\    

   			\quad FedAVE\cite{wang2024fedave}     & \multicolumn{1}{c}{46.43$\pm$0.6}          & \multicolumn{1}{c}{40.99$\pm$0.2}          & \multicolumn{1}{c}{46.64$\pm$0.6}              & \multicolumn{1}{c}{48.34$\pm$0.5}              & \multicolumn{1}{c|}{48.13$\pm$0.6}                     & \multicolumn{1}{c}{46.58$\pm$0.4}          & \multicolumn{1}{c|}{46.73$\pm$0.3}          & \multicolumn{1}{c}{47.29$\pm$0.8}  & \multicolumn{1}{c|}{49.21$\pm$0.6}   & \multicolumn{1}{c}{\underline{87.44}$\pm$0.1} & \multicolumn{1}{c}{\underline{84.79}$\pm$0.5} & \multicolumn{1}{c}{79.53$\pm$0.9} & \multicolumn{1}{c}{84.47$\pm$0.4} & \multicolumn{1}{c}{86.83$\pm$0.1}    \\         
			
			\quad Ours   & \multicolumn{1}{c}{\textbf{48.60}$\pm$0.7}          & \multicolumn{1}{c}{\textbf{43.39}$\pm$0.2}          & \multicolumn{1}{c}{\textbf{49.41}$\pm$0.1}              & \multicolumn{1}{c}{\textbf{49.09}$\pm$0.1}              & \multicolumn{1}{c|}{\textbf{50.48}$\pm$0.1}                     & \multicolumn{1}{c}{\textbf{48.64}$\pm$0.3}          & \multicolumn{1}{c|}{\textbf{49.68}$\pm$0.4}          & \multicolumn{1}{c}{\textbf{49.23}$\pm$0.1}  & \multicolumn{1}{c|}{\textbf{49.34}$\pm$0.2}   & \multicolumn{1}{c}{\textbf{87.60}$\pm$0.1} & \multicolumn{1}{c}{\textbf{84.99}$\pm$0.3} & \multicolumn{1}{c}{\textbf{87.57}$\pm$0.6} & \multicolumn{1}{c}{\textbf{88.08}$\pm$0.3} & \multicolumn{1}{c}{\textbf{88.17}$\pm$0.3}    \\       
			\bottomrule
		\end{tabular}
	}
	\renewcommand\thetable{6}
    \caption{~\textbf{Comparison results of the maximum test accuracy ($\%$)} with state-of-the-art methods on CIFAR10 and Fashion MNIST. The reported results are averaged over 5 runs with different random seeds. (A higher value indicates better accuracy.)}
	\label{tab6}
\end{table*}



\section*{B. Final Rewards of Clients}
Figure \ref{figure 6.} shows the distributions of contributions and the final rewards of clients under different scenes (i.e., 
POW, CLA, DIR (2.0) in CIFAR10 and SVHN) by FedSAC and the compared methods. Existing methods allocate rewards to clients lack sufficient differentiation, resulting in an ongoing unfairness for high-contributing clients. For example in SVHN-CLA (Figure \ref{figure 6.} (d)), the contributions (i.e., Standalone) of $Client_3$ and $Client_{10}$ differ significantly. However, CFFL and CGSV do not exhibit a substantial difference in the rewards assigned to them. In addition, FedSAC effectively differentiates the rewards it received, thereby ensuring the collaborative fairness in FL.

\begin{table}
	\centering
	\resizebox{1.0\linewidth}{!}{
  \renewcommand\arraystretch{1.3}
		\large
		\begin{tabular}{@{}l|c|cccc@{}}
			\toprule
			\quad Scene & \multicolumn{1}{c}{POW}                     & \multicolumn{1}{c}{DIR(1.0)}               & \multicolumn{1}{c}{DIR(2.0)}            & \multicolumn{1}{c}{DIR(3.0)}  \\
			\midrule   
			
			\quad 		$FedSAC(\beta=2)$& \multicolumn{1}{c}{\textbf{49.23(44.77)}}         & \multicolumn{1}{c}{\textbf{49.66(39.53)}}              & \multicolumn{1}{c}{\textbf{50.60(37.87)}}              & \multicolumn{1}{c}{\textbf{51.37(43.19)}}     \\               
			
			\quad 	$FedSAC(\beta=5)$& \multicolumn{1}{c}{48.43(37.73)}          & \multicolumn{1}{c}{49.10(35.76)}              & \multicolumn{1}{c}{50.01(30.91)}              & 
			\multicolumn{1}{c}{49.85(38.46)}    \\    
			
			\quad 	$FedSAC(\beta=10)$& \multicolumn{1}{c}{48.61(30.91)}        & \multicolumn{1}{c}{49.06(31.36)}              & \multicolumn{1}{c}{47.08(23.72)}              & 
			\multicolumn{1}{c}{49.04(28.54)}    \\

			\quad 		$FedSAC(\beta=20)$& \multicolumn{1}{c}{46.53(24.10)}    & \multicolumn{1}{c}{46.98(28.41)}              & \multicolumn{1}{c}{43.50(20.17)}              & \multicolumn{1}{c}{46.03(25.70)}    \\    
			
			\bottomrule
		\end{tabular}
	}
	\renewcommand\thetable{7}
	\caption{The maximum test accuracy (\%) achieved by FedSAC across different $\beta$, given a fairness threshold of $\rho > 95\%$, on CIFAR10. Values in the middle brackets represent the minimum test accuracy (\%) among 10 clients.}
	\label{table7}
\end{table}

\begin{table*}[ht]
	\centering
	\resizebox{1.0\linewidth}{!}{
		\large
		\begin{tabular}{@{}l|c|ccccc|ccccc|ccccc@{}}
			\toprule
			\quad  Dataset        & \multicolumn{5}{c|}{CIFAR10} &\multicolumn{5}{c|}{SVHN} &\multicolumn{5}{c}{Fashion MNIST}\\ 
			\midrule
			\quad  No. Clients        & \multicolumn{5}{c|}{10} &\multicolumn{5}{c|}{10} &\multicolumn{5}{c}{10}\\ 
			\midrule
			\quad Scene & \multicolumn{1}{c}{POW}            & \multicolumn{1}{c}{CLA}                & \multicolumn{1}{c}{DIR(1.0)}               & \multicolumn{1}{c}{DIR(2.0)}            & \multicolumn{1}{c|}{DIR(3.0)}     
			& \multicolumn{1}{c}{POW}            & \multicolumn{1}{c}{CLA}                  & \multicolumn{1}{c}{DIR(1.0)}               & \multicolumn{1}{c}{DIR(2.0)}            & \multicolumn{1}{c|}{DIR(3.0)} & \multicolumn{1}{c}{POW}            & \multicolumn{1}{c}{CLA}                  & \multicolumn{1}{c}{DIR(1.0)}               & \multicolumn{1}{c}{DIR(2.0)}            & \multicolumn{1}{c}{DIR(3.0)}\\
			\midrule   
			\quad 	FedAvg\cite{Mcmahan2017}          & \multicolumn{1}{c}{26.27}          & \multicolumn{1}{c}{26.27}          & \multicolumn{1}{c}{26.27}              & \multicolumn{1}{c}{26.27}              & \multicolumn{1}{c|}{26.27}              & 
			\multicolumn{1}{c}{26.27}                         & \multicolumn{1}{c}{26.27}              & \multicolumn{1}{c}{26.27}              & \multicolumn{1}{c}{26.27}              & 
			\multicolumn{1}{c|}{26.27}   & \multicolumn{1}{c}{7.97}          & \multicolumn{1}{c}{7.97}          & \multicolumn{1}{c}{7.97}                        & \multicolumn{1}{c}{7.97}              & 
			\multicolumn{1}{c}{7.97} \\               
			
			\quad q-FFL\cite{li2019fair}           & \multicolumn{1}{c}{26.27}          & \multicolumn{1}{c}{26.27}          & \multicolumn{1}{c}{26.27}              & \multicolumn{1}{c}{26.27}              & \multicolumn{1}{c|}{26.27}              & 
			\multicolumn{1}{c}{26.27}                         & \multicolumn{1}{c}{26.27}              & \multicolumn{1}{c}{26.27}              & \multicolumn{1}{c}{26.27}              & 
			\multicolumn{1}{c|}{26.27}   & \multicolumn{1}{c}{7.97}          & \multicolumn{1}{c}{7.97}          & \multicolumn{1}{c}{7.97}                        & \multicolumn{1}{c}{7.97}              & 
			\multicolumn{1}{c}{7.97} \\    
			
			\quad CFFL\cite{lyu2020collaborative}         & \multicolumn{1}{c}{26.27}          & \multicolumn{1}{c}{26.27}          & \multicolumn{1}{c}{26.27}              & \multicolumn{1}{c}{26.27}              & \multicolumn{1}{c|}{26.27}              & 
			\multicolumn{1}{c}{26.27}                         & \multicolumn{1}{c}{26.27}              & \multicolumn{1}{c}{26.27}              & \multicolumn{1}{c}{26.27}              & 
			\multicolumn{1}{c|}{26.27}   & \multicolumn{1}{c}{7.97}          & \multicolumn{1}{c}{7.97}          & \multicolumn{1}{c}{7.97}                        & \multicolumn{1}{c}{7.97}              & 
			\multicolumn{1}{c}{7.97} \\    
			
			\quad CGSV\cite{xu2021gradient}           & \multicolumn{1}{c}{26.27}          & \multicolumn{1}{c}{26.27}          & \multicolumn{1}{c}{26.27}              & \multicolumn{1}{c}{26.27}              & \multicolumn{1}{c|}{26.27}              & 
			\multicolumn{1}{c}{26.27}                         & \multicolumn{1}{c}{26.27}              & \multicolumn{1}{c}{26.27}              & \multicolumn{1}{c}{26.27}              & 
			\multicolumn{1}{c|}{26.27}   & \multicolumn{1}{c}{7.97}          & \multicolumn{1}{c}{7.97}          & \multicolumn{1}{c}{7.97}                        & \multicolumn{1}{c}{7.97}              & 
			\multicolumn{1}{c}{7.97} \\

   			\quad FedAVE\cite{wang2024fedave}           & \multicolumn{1}{c}{26.27}          & \multicolumn{1}{c}{26.27}          & \multicolumn{1}{c}{26.27}              & \multicolumn{1}{c}{26.27}              & \multicolumn{1}{c|}{26.27}              & 
			\multicolumn{1}{c}{26.27}                         & \multicolumn{1}{c}{26.27}              & \multicolumn{1}{c}{26.27}              & \multicolumn{1}{c}{26.27}              & 
			\multicolumn{1}{c|}{26.27}   & \multicolumn{1}{c}{7.97}          & \multicolumn{1}{c}{7.97}          & \multicolumn{1}{c}{7.97}                        & \multicolumn{1}{c}{7.97}              & 
			\multicolumn{1}{c}{7.97} \\    
			
			\quad Ours      & \multicolumn{1}{c}{\textbf{23.57}}          & \multicolumn{1}{c}{\textbf{23.58}}              & \multicolumn{1}{c}{\textbf{23.39}}              & \multicolumn{1}{c}{\textbf{24.02}}              & 
			\multicolumn{1}{c|}{\textbf{23.63}}   & \multicolumn{1}{c}{\textbf{20.86}}          & \multicolumn{1}{c}{\textbf{22.31}}                    & \multicolumn{1}{c}{\textbf{23.85}}              & \multicolumn{1}{c}{\textbf{15.51}}              & 
			\multicolumn{1}{c|}{\textbf{21.75}}   & \multicolumn{1}{c}{\textbf{6.89}}          & \multicolumn{1}{c}{\textbf{7.28}}               & \multicolumn{1}{c}{\textbf{5.30}}              & \multicolumn{1}{c}{\textbf{4.77}}              &  \multicolumn{1}{c}{\textbf{5.23}} \\
			\bottomrule
		\end{tabular}
	}
	\renewcommand\thetable{8}
	\caption{Comparison on communication costs (MB per round) of FedSAC and the baselines framework.}
	\label{table8}
\end{table*}

\begin{table*}[ht]
	\centering
	\resizebox{1.0\linewidth}{!}{
		\large
		\begin{tabular}{@{}l|c|ccccc|ccccc|ccccc@{}}
			\toprule
			\quad  Dataset        & \multicolumn{5}{c|}{CIFAR10} &\multicolumn{5}{c|}{SVHN} &\multicolumn{5}{c}{Fashion MNIST}\\ 
			\midrule
			\quad  No. Clients        & \multicolumn{5}{c|}{10} &\multicolumn{5}{c|}{10} &\multicolumn{5}{c}{10}\\ 
			\midrule
			\quad Scene & \multicolumn{1}{c}{POW}            & \multicolumn{1}{c}{CLA}                & \multicolumn{1}{c}{DIR(1.0)}               & \multicolumn{1}{c}{DIR(2.0)}            & \multicolumn{1}{c|}{DIR(3.0)}     
			& \multicolumn{1}{c}{POW}            & \multicolumn{1}{c}{CLA}                  & \multicolumn{1}{c}{DIR(1.0)}               & \multicolumn{1}{c}{DIR(2.0)}            & \multicolumn{1}{c|}{DIR(3.0)} & \multicolumn{1}{c}{POW}            & \multicolumn{1}{c}{CLA}                  & \multicolumn{1}{c}{DIR(1.0)}               & \multicolumn{1}{c}{DIR(2.0)}            & \multicolumn{1}{c}{DIR(3.0)}\\
			\midrule   
			
			\quad 	$\eta$       & \multicolumn{1}{c}{0.05}          & \multicolumn{1}{c}{0.05}      & \multicolumn{1}{c}{0.05}              & \multicolumn{1}{c}{0.03}              & 
			\multicolumn{1}{c|}{0.03}   & \multicolumn{1}{c}{0.03}          & \multicolumn{1}{c}{0.05}       & \multicolumn{1}{c}{0.05}          & \multicolumn{1}{c}{0.05}              & \multicolumn{1}{c|}{0.05}              & \multicolumn{1}{c}{0.1}              & 
			\multicolumn{1}{c}{0.05}       & \multicolumn{1}{c}{0.05}          & \multicolumn{1}{c}{0.05}        & \multicolumn{1}{c}{0.05}  \\               
			
			\quad $E$          & \multicolumn{1}{c}{15}          & \multicolumn{1}{c}{20}     & \multicolumn{1}{c}{20}              & \multicolumn{1}{c}{20}              & 
			\multicolumn{1}{c|}{20}   & \multicolumn{1}{c}{20}          & \multicolumn{1}{c}{20}          & \multicolumn{1}{c}{20}               & \multicolumn{1}{c}{20}              & \multicolumn{1}{c|}{20}              & \multicolumn{1}{c}{20}              & 
			\multicolumn{1}{c}{20}    & \multicolumn{1}{c}{20}          & \multicolumn{1}{c}{20}   & \multicolumn{1}{c}{15}  \\    
			
			\quad 	$\beta$      & \multicolumn{1}{c}{5}          & \multicolumn{1}{c}{3}       & \multicolumn{1}{c}{10}              & \multicolumn{1}{c}{2}              & 
			\multicolumn{1}{c|}{10}   & \multicolumn{1}{c}{5}          & \multicolumn{1}{c}{3}  & \multicolumn{1}{c}{5}          & \multicolumn{1}{c}{10}              & \multicolumn{1}{c|}{5}              & \multicolumn{1}{c}{10}              & 
			\multicolumn{1}{c}{1} & \multicolumn{1}{c}{10}          & \multicolumn{1}{c}{20}    & \multicolumn{1}{c}{25} \\    
			
			\bottomrule
		\end{tabular}
	}
	\renewcommand\thetable{9}
	\caption{The optimal hyperparameters in given scenarios. Learning rate $\eta$, the number of local update steps $E$, hyper-parameter $\beta$. Batch sizes $B$ are set as 32, 64, and 32 for CIFAR10, SVHN, and FashionMNIST, respectively.}
	\label{tab9}
\end{table*}

\section*{C. Varying Numbers of Clients}
To verify the effectiveness of FedSAC in scenarios with varying numbers of clients, we conduct experiments by increasing the number of local clients to 20, 40, and 60, respectively. Table~\ref{tab5} presents the fairness results achieved by FedSAC and the compared methods, while Table~\ref{tab6} shows the maximum local model performance achieved by these methods. In the large client number settings (No. Clients = 20, 40, and 60), FedSAC outperforms all baseline methods in terms of fairness (refer to Table~\ref{tab5}) and accuracy (refer to Table~\ref{tab6}). 
The results demonstrate that FedSAC can effectively implement bounded collaborative fairness in scenarios with varying numbers of clients.


\section*{D. The Impact of $\beta$ on The Experiment}
In Table \ref{table7}, we present the performance of FedSAC with different values of $\beta$ on each scene of CIFAR10. The experiments demonstrate that as $\beta$ increases, the maximum test accuracy will gradually decrease. This is because the size of the submodels downloaded by clients increases as $\beta$ decreases. When $\beta$ is small, the submodels of low-contribution clients contain more neurons, enabling effective training to enhance all local model performance.

\section*{E. The Communication Cost Experiments}
The most works on collaborative fairness require full clients’ information~\cite{lyu2020collaborative,xu2020reputation,xu2021gradient}, which will inevitably introduce large communication overhead and computation cost. In addition, this communication and computation overhead of full sampling problem is tolerable for most cases in cross-silo FL scenarios ~\cite{du2022flamby}, such as healthcare and finance, because there are only dozens of clients in cross-silo FL scenarios.  In the Table \ref{table8}, the results show that our FedSAC demonstrates less communication costs than all the baseline methods in three datasets across all settings. In addition, our FedSAC does not introduce additional communication, because the evaluation of neuron importance is conducted on the server without communicating with clients, effectively preventing any communication overhead. In the Table \ref{table8}, FedSAC exhibits a communication complexity of O(d*m) per round~\cite{horvath2021fjord}, where m$<$=1 represents the average ratio of the parameters of the submodel compared to the global model. In conclusion, all baseline methods show a higher communication complexity of O(d) than FedSAC.

\section*{F. Details on Hyper-parameters}
For each dataset, the local data of each client was partitioned into training and validation sets. Then, we tuned each dataset hyper-parameters by using grid search with FedAvg. Subsequently, we applied the optimal parameters obtained from the validation dataset. The optimized hyper-parameters for scenarios (i.e., POW, CLA, DIR(1.0), DIR(2.0), and DIR(3.0)) are shown in Table \ref{tab9}.

\end{document}